\documentclass[12pt]{article}
\usepackage[utf8]{inputenc}
\usepackage{hyperref}
\usepackage{amsmath}
\usepackage{graphicx}
\usepackage{enumitem}
\usepackage{tcolorbox}
\usepackage{algorithm}
\usepackage{algorithmic}
\usepackage{graphicx}
\usepackage{subcaption}
\usepackage{amsmath}
\usepackage{amssymb}
\usepackage{amsfonts}
\usepackage{float}

\title{\vspace{-2cm}Generative AI for Vision: A Comprehensive Study of Frameworks and Applications\vspace{1cm}}

\author{
    Fouad Bousetouane\textsuperscript{1,2} \\[1em] 
    \textsuperscript{1}\text{The University of Chicago, USA} \\[0.75em] 
    \textsuperscript{2}\text{2ndsight.ai} \\[1.5em] 
    {\small \href{mailto:bousetouane@uchicago.edu}{\texttt{bousetouane@uchicago.edu}}}
}

\date{}

\begin{document}

\maketitle

\begin{abstract}
Generative AI is transforming image synthesis, enabling the creation of high-quality, diverse, and photorealistic visuals across industries like design, media, healthcare, and autonomous systems. Advances in techniques such as image-to-image translation, text-to-image generation, domain transfer, and multimodal alignment have broadened the scope of automated visual content creation, supporting a wide spectrum of applications. These advancements are driven by models like Generative Adversarial Networks (GANs), conditional frameworks, and diffusion-based approaches such as Stable Diffusion.

This work presents a structured classification of image generation techniques based on the nature of the input, organizing methods by input modalities like noisy vectors, latent representations, and conditional inputs. We explore the principles behind these models, highlight key frameworks including DALL-E, ControlNet, and DeepSeek Janus-Pro, and address challenges such as computational costs, data biases, and output alignment with user intent. By offering this input-centric perspective, this study bridges technical depth with practical insights, providing researchers and practitioners with a comprehensive resource to harness generative AI for real-world applications.
\end{abstract}

\newpage
\tableofcontents
\newpage

\section{Introduction}

Generative AI is revolutionizing the field of computer vision, bringing forth transformative methods that empower machines to generate, understand, and manipulate visual content with unprecedented sophistication. While recent breakthroughs in large language models (LLMs) such as GPT \cite{openai2024gpt4technicalreport}, LLama \cite{touvron2023llama2openfoundation}, and DeepSeek \cite{deepseek-llm} have dominated the text domain, the vision domain presents unique challenges due to its inherently complex, high-dimensional, and infinitely variable nature. These challenges require advanced computational models capable of navigating and interpreting visual data's rich complexity, far beyond the structured confines of linguistic constructs.

Recent advancements in generative models, including Generative Adversarial Networks (GANs), diffusion models, and conditional frameworks, have redefined the boundaries of what is possible in computer vision. From visual synthesis to domain adaptation, these technologies are enabling novel applications across industries. For instance, text-to-image generation models, such as OpenAI's DALL-E \cite{ramesh2022openai} and Google’s Imagen \cite{saharia2022}, translate natural language descriptions into highly detailed and contextually appropriate images, revolutionizing fields like design, marketing, and entertainment by facilitating scalable and tailored content creation.

Similarly, image-to-image translation models, such as CycleGAN \cite{CycleGAN2017}, enable seamless transformations between visual domains, supporting tasks like artistic style transfer and medical image enhancement. These advancements extend to domain transfer techniques, which adapt images across distinct contexts while preserving semantic integrity—an approach invaluable for scientific visualization, simulation, and autonomous systems.

The integration of vision and language has further expanded the scope of generative AI. Models like CLIP \cite{CLIP} bridge image-text alignment, enabling applications such as multimodal search, image captioning, and interactive AI systems. This synergy between vision and language demonstrates the growing importance of multimodal frameworks in building intelligent systems capable of understanding and interacting with diverse data types.
\\
\\
Generative AI’s influence in vision spans a range of impactful applications:

\begin{itemize}
    \item \textbf{E-commerce:} Generative models enhance customer engagement by dynamically creating personalized product visuals and immersive shopping experiences.
    \item \textbf{Scientific Research:} Tools like Stable Diffusion \cite{StableDif} generate high-resolution depictions of complex phenomena, such as molecular structures and astrophysical simulations, accelerating innovation in fields like biology and physics.
    \item \textbf{Autonomous Systems:} Generative AI simulates realistic environments for training and testing, bridging domain gaps and reducing development costs for self-driving vehicles and robotic systems.
\end{itemize}

Generative AI in vision holds great potential but faces several challenges. It requires significant computational resources to handle complex visual data and relies on advanced architectures. Ensuring outputs match user intent is difficult, especially with biases in training data that can affect fairness. Overcoming these issues is essential to make generative AI more inclusive and equitable.
\\ 

This article provides a comprehensive overview of generative AI in computer vision, covering methodologies, applications, and challenges. Section \ref{Techniques} categorizes image generation techniques by input type, highlighting diverse approaches. Section \ref{Noisy} examines noisy vector-based methods, such as GANs and diffusion models, while Section \ref{VAEs} explores latent space models like Variational Autoencoders (VAEs). Section \ref{prompt} focuses on text-to-image generation frameworks, including DALL-E, Stable Diffusion and DeepSeek Janus-Pro, and Section \ref{Conditional} reviews conditional input methods like ControlNet for tailored outputs. Section \ref{ethical} discusses challenges such as biases, computational costs, and ethical considerations. Finally, Section \ref{conclusion} summarizes key insights and outlines future directions, including multimodal advancements and agentic AI for vision applications.

\section{Image Generation Techniques}
\label{Techniques}
Image generation techniques have evolved significantly, leveraging various forms of input to produce compelling and diverse outputs. The type of input plays a critical role in shaping the design, functionality, and applications of generative models, influencing their ability to address specific challenges and use cases. In this work, we propose a categorization of image generation techniques based on the nature of the input, offering a structured framework to understand their methodologies, capabilities, and real-world applications.

\begin{figure}[h]
    \centering
    \includegraphics[width=1\textwidth]{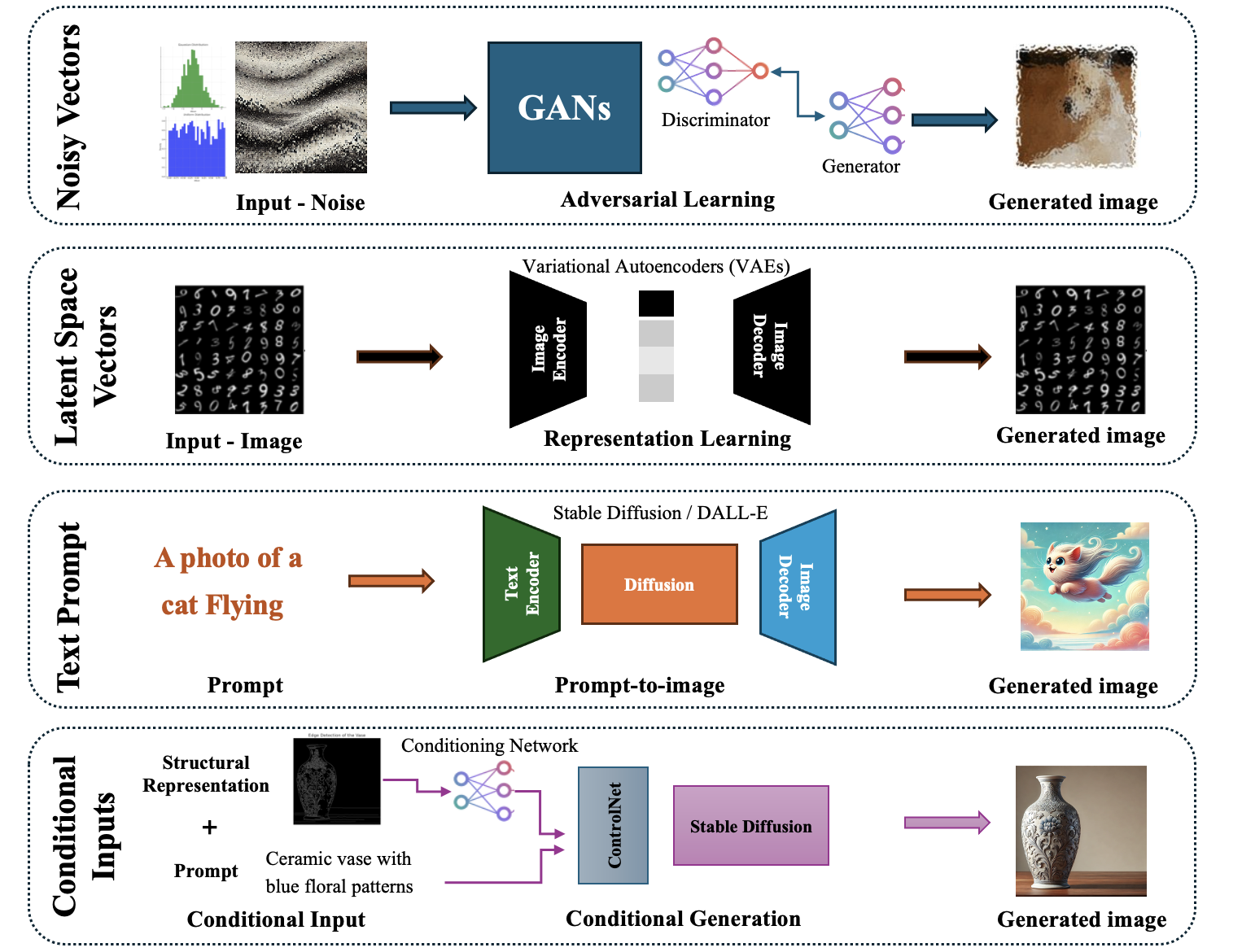} 
    \caption{Key categories of input-driven image generation techniques. These categories illustrate the distinct workflows and methodologies involved in generating compelling visual outputs}
    \label{fig:image_generation_categories}
\end{figure}

As illustrated in Figure~\ref{fig:image_generation_categories}, the input-driven categories in image generation demonstrate the diverse workflows used to produce meaningful outputs, spanning from random noise to structured textual prompts. This categorization provides a foundational structure for developing state-of-the-art generative models, supporting tailored applications across a variety of domains.

The key categories of the proposed classification of input-driven image generation techniques are as follows:

\begin{itemize}
    \item \textbf{Noisy Vectors:} Methods like Generative Adversarial Networks (GANs) use random noise as input to generate diverse and realistic images through adversarial training.
\newpage
    
    \item \textbf{Latent Space Representations:} Techniques like Variational Autoencoders (VAEs) encode input data into compressed embeddings and decode them to reconstruct or generate new images, facilitating structured manipulation of the latent space.
    \item \textbf{Conditional Inputs:} Conditional models rely on specific inputs, such as labels, sketches, or structural data, to guide the generation process and produce tailored outputs aligned with user-defined requirements.
    \item \textbf{Textual Descriptions:} Text prompts serve as inputs for text-to-image generation models, such as DALL-E \cite{ramesh2022openai} and Stable Diffusion \cite{rombach2022high}, transforming natural language prompts into visual content, seamlessly integrating linguistic and visual modalities.
\end{itemize}

This categorization provides a comprehensive understanding of the diverse approaches used in image generation, offering insights into their potential applications across  industries. By leveraging these techniques, researchers and practitioners can address complex challenges, create tailored visual content, and explore the limitless possibilities of generative AI in computer vision.

In the next sections, we explore these categories in depth, detailing the underlying models, their key features, and real-world use cases that highlight their practical significance.

\section{Noisy Vectors: GANs \& Diffusion Models}
\label{Noisy}
Noisy vector-based models rely on random noise as input to generate realistic and diverse images. This category includes two primary approaches: GANs and diffusion models.

GANs \cite{goodfellow2014generative} use an adversarial framework, where a generator creates images from random noise, and a discriminator evaluates their realism. This competition drives the generator to produce high-quality outputs, making GANs ideal for tasks requiring sharp, realistic visuals \cite{arjovsky2017wasserstein, brock2018large}.

Diffusion models \cite{ho2020denoising} refine noisy inputs through an iterative denoising process, learning to reverse the noise addition. They excel in generating diverse and high-fidelity images, overcoming challenges like mode collapse common in GANs \cite{dhariwal2021diffusion}.
The next sections explore their architectures, training methods, and applications in depth.

\subsection{Generative Adversarial Networks (GANs)}
GANs are foundational to the field of generative AI, beginning with the Vanilla GAN \cite{goodfellow2014generative}, which introduced the adversarial training framework. This setup involves a generator and discriminator competing to improve their performance iteratively, enabling the generation of realistic images from random noise. The simplicity and adaptability of the Vanilla GAN have inspired a multitude of specialized architectures tailored to various tasks and applications.

Prominent types include CycleGAN \cite{CycleGAN2017} for unpaired image-to-image translation, Pix2Pix \cite{isola2017image} for paired image-to-image translation, StyleGAN \cite{karras2019style} for high-quality image synthesis with fine-grained control over style, and Super-Resolution GAN (SRGAN) \cite{ledig2017photo} for enhancing low-resolution images into high-resolution outputs. Each type introduces unique innovations to address specific challenges in image generation and processing.

\subsubsection{Vanilla GANs}

Vanilla GANs are the foundational architecture for Generative Adversarial Networks (GANs), introduced by Ian Goodfellow et al. \cite{goodfellow2014generative}. The primary objective of a Vanilla GAN is to train two neural networks—a generator and a discriminator—in an adversarial framework. The generator learns to produce realistic images from random noise, while the discriminator learns to distinguish real images from generated ones. Together, they play a minimax game defined by the following objective function:

\begin{equation}
\min_G \max_D \mathbb{E}_{x \sim p_{\text{data}}(x)}[\log D(x)] + \mathbb{E}_{z \sim p_z(z)}[\log(1 - D(G(z)))]
\end{equation}

\paragraph{Objective Function Components}
\begin{itemize}
    \item \textbf{Generator ($G$):} Maps a noise vector $z$ from a latent space to a generated image $G(z)$.
    \item \textbf{Discriminator ($D$):} Outputs a probability indicating whether the input is a real image ($x$) or a generated image ($G(z)$).
    \item \textbf{Real Image Distribution ($p_{\text{data}}(x)$):} Represents the distribution of real images in the dataset.
    \item \textbf{Noise Distribution ($p_z(z)$):} Defines the distribution of noise vectors (e.g., Gaussian or uniform) used as input for the generator.
    \item \textbf{Log Probabilities:} $\log D(x)$ measures the discriminator's ability to identify real images, while $\log(1 - D(G(z)))$ evaluates its accuracy in identifying fake images.
\end{itemize}

\begin{figure}[h]
    \centering
    \includegraphics[width=1\textwidth]{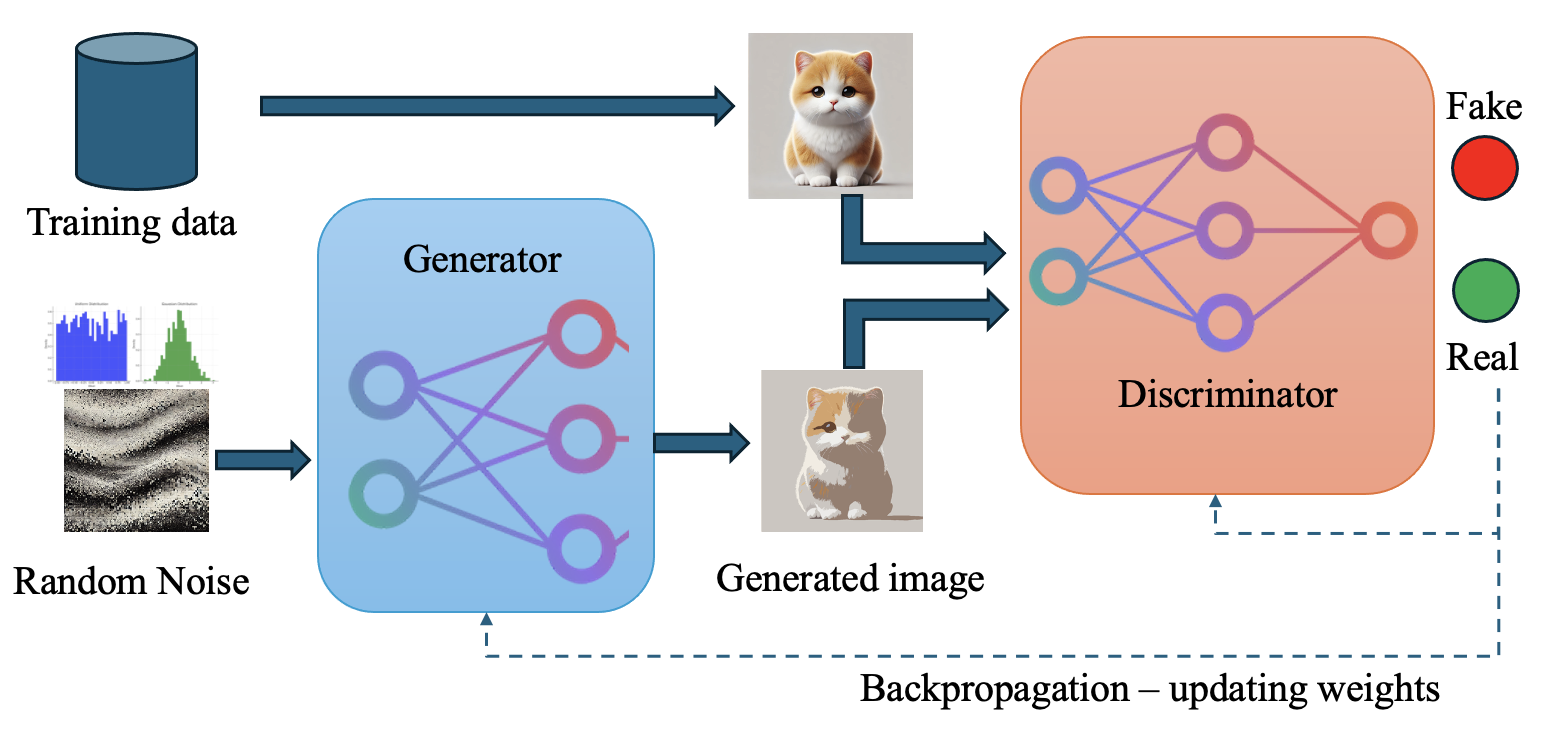} 
    \caption{Adversarial training setup for a Vanilla GAN. The generator ($G$) maps noise vectors to generated images, while the discriminator ($D$) evaluates whether inputs are real or generated. The adversarial process ensures that both networks improve iteratively.}
    \label{fig:adversarial_training}
\end{figure}

This adversarial training framework ensures that the generator improves in creating realistic images while the discriminator refines its ability to distinguish between real and fake data. Figure~\ref{fig:adversarial_training} illustrates the adversarial training setup, showcasing the interaction between the generator and discriminator.

\paragraph{Practical Applications of Vanilla GANs:} Vanilla GANs have demonstrated their versatility in various fields. Below are examples of popular use cases:

\begin{enumerate}
    \item \textbf{Data Augmentation for Training Deep Vision Models}
    \begin{itemize}
        \item \textbf{Use Case:} Expanding training datasets for computer vision models in scenarios with limited labeled data, such as rare medical images or unique object categories.
        \item \textbf{Process:} Vanilla GANs synthesize realistic images to augment datasets, ensuring better representation of rare or underrepresented categories.
        \item \textbf{Impact:} Increases dataset diversity, enhances model robustness, and improves generalization for vision tasks like object detection or segmentation.
    \end{itemize}

    \item \textbf{Gaming and Virtual Reality (VR)}
    \begin{itemize}
        \item \textbf{Use Case:} Designing realistic textures, characters, and environments for immersive gaming and VR experiences.
        \item \textbf{Process:} Vanilla GANs generate high-quality visual assets, including lifelike textures, dynamic landscapes, and character models, enriching the visual quality of virtual worlds.
        \item \textbf{Impact:} Boosts realism and engagement, making gaming and VR environments more compelling and interactive.
    \end{itemize}

    \item \textbf{Surveillance and Anomaly Detection}
    \begin{itemize}
        \item \textbf{Use Case:} Enhancing training datasets for surveillance systems by generating synthetic data, such as rare security breach scenarios or unusual object appearances.
        \item \textbf{Process:} Vanilla GANs create realistic images of anomalies, such as tampered objects or unexpected items, to improve the ability of vision models to detect anomalies.
        \item \textbf{Impact:} Enhances the accuracy and reliability of anomaly detection in security and monitoring systems, reducing false positives and improving response times.
    \end{itemize}
\end{enumerate}

\subsubsection{Pix2Pix: Paired Image Translation}

Pix2Pix \cite{isola2017image} is a GAN-based framework tailored for paired image-to-image translation tasks, where each input image has a corresponding target image. By enabling precise mappings between domains, this framework addresses industry challenges such as anomaly detection, image enhancement, and automated mapping. Its conditional GAN structure, leveraging paired datasets, ensures outputs are both realistic and semantically aligned with the target data.

\begin{figure}[h]
    \centering
    \includegraphics[width=1\textwidth]{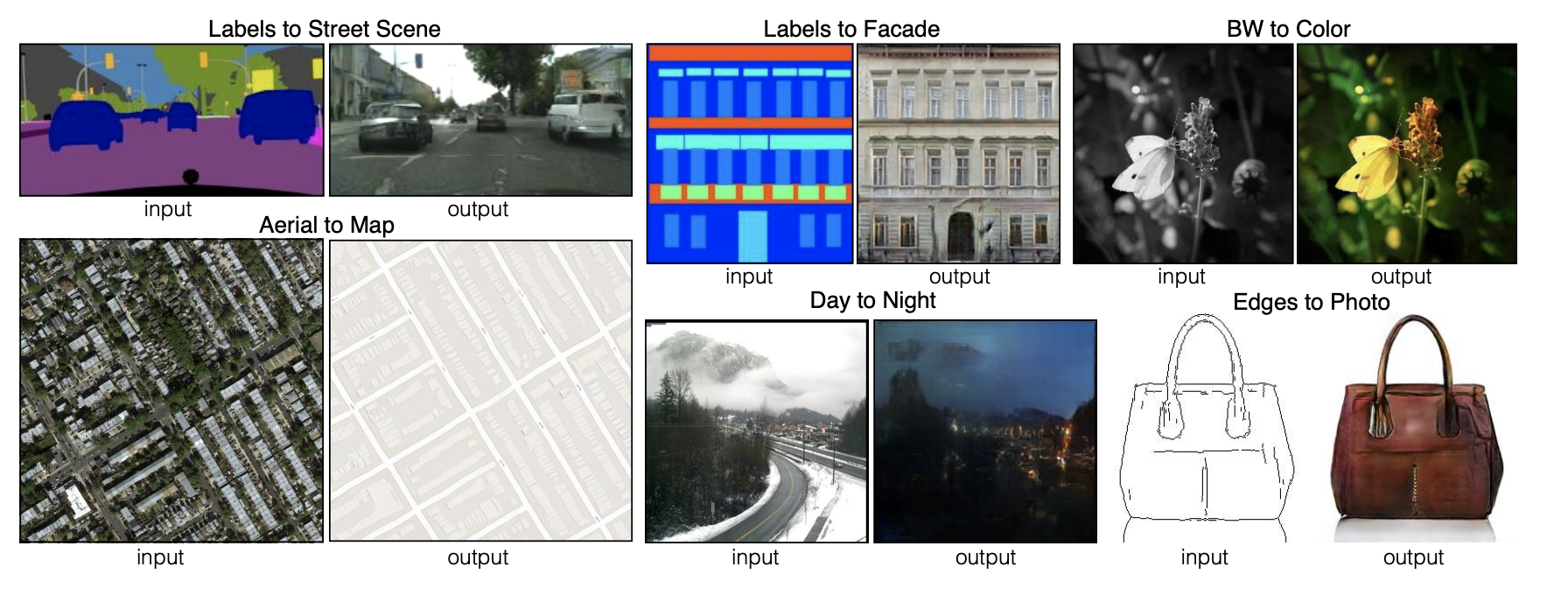}
    \caption{Pix2Pix supports diverse paired image-to-image translation tasks, such as labels-to-street scenes, aerial images to maps, black-and-white to color, and edges to photos. These examples highlight its versatility in addressing various vision challenges \cite{isola2017image}.}
    \label{fig:pix2pix_tasks}
\end{figure}

Figure~\ref{fig:pix2pix_tasks} illustrates the versatility of Pix2Pix in handling diverse tasks, ranging from practical applications like urban planning to creative endeavors such as artistic rendering.

\paragraph{Key Features of Pix2Pix}
\begin{itemize}
    \item \textbf{Paired Training:} Leverages paired datasets to ensure precise and reliable mappings between input and output domains.
    \item \textbf{Conditional GAN Framework:} The generator produces outputs conditioned on input images, while the discriminator evaluates their authenticity and realism.
    \item \textbf{Combination of Losses:} Adversarial loss promotes realism, and L1 loss ensures the outputs are closely aligned with the target images.
\end{itemize}

\begin{figure}[h]
    \centering
    \includegraphics[width=1\textwidth]{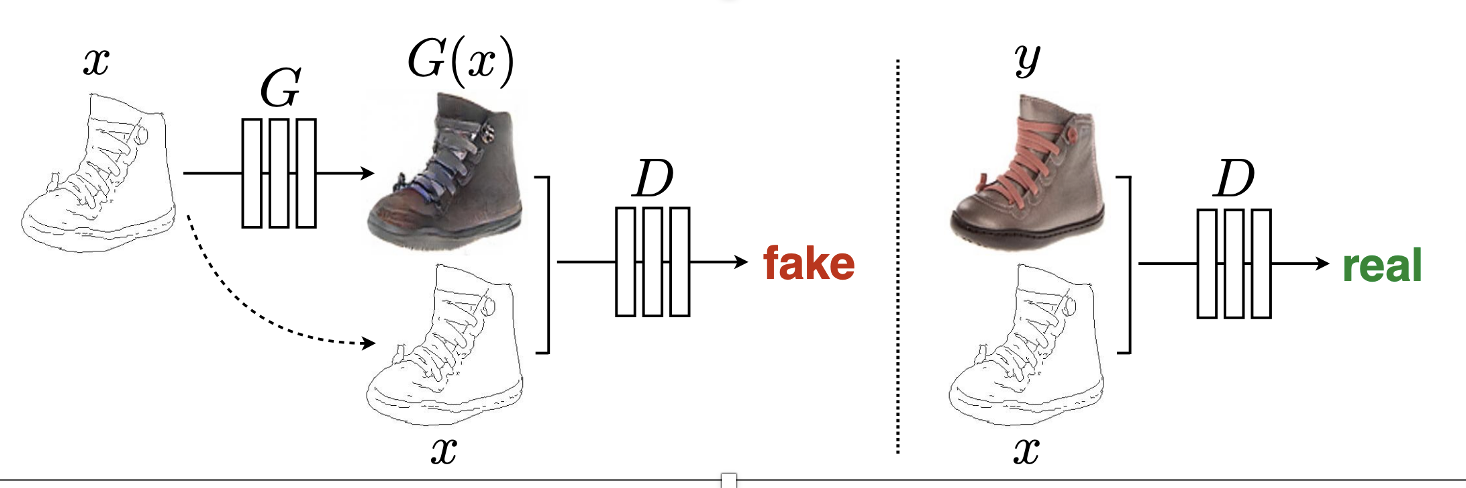}
    \caption{Training process where the generator ($G$) and discriminator ($D$) observe the input edge map to classify real and fake pairs \cite{isola2017image}.}
    \label{fig:pix2pix_training}
\end{figure}

\paragraph{How Pix2Pix Works (Simplified):} Pix2Pix transforms input images into target outputs using a generator and discriminator in tandem. The generator synthesizes outputs conditioned on the input, while the discriminator evaluates the pair for authenticity. Combining adversarial and L1 losses ensures visually realistic and semantically accurate results, making Pix2Pix a robust tool for various industries. Figure~\ref{fig:pix2pix_training} illustrates this process in the context of edge-to-photo translation.

\paragraph{Practical Applications of Pix2Pix}
\begin{enumerate}
    \item \textbf{Oil and Gas: Pipeline Inspection}
    \begin{itemize}
        \item \textbf{Use Case:} Enhance pipeline anomaly detection from low-quality inspection images.
        \item \textbf{Process:} Pix2Pix is trained on paired datasets of noisy and enhanced pipeline images to generate clearer visuals with highlighted defects.
        \item \textbf{Impact:} Improves defect detection reliability and accelerates maintenance, reducing downtime and costs.
    \end{itemize}

    \item \textbf{Geospatial Analysis: Automated Mapping}
    \begin{itemize}
        \item \textbf{Use Case:} Convert aerial images into detailed, segmented maps for urban planning and resource management.
        \item \textbf{Process:} Pix2Pix translates raw aerial data into labeled maps by learning the mapping from unprocessed to segmented imagery.
        \item \textbf{Impact:} Streamlines terrain analysis, aiding efficient urban planning and resource allocation.
    \end{itemize}

    \item \textbf{Healthcare: Medical Imaging Enhancement}
    \begin{itemize}
        \item \textbf{Use Case:} Improve low-quality scans for better anomaly detection in medical diagnostics.
        \item \textbf{Process:} Pix2Pix generates high-resolution outputs by refining noisy medical images, enhancing visual clarity and diagnostic precision.
        \item \textbf{Impact:} Assists in timely, accurate diagnoses, improving patient outcomes and treatment planning.
    \end{itemize}
\end{enumerate}

\paragraph{Implementation Framework:} The \texttt{pytorch-CycleGAN-and-pix2pix} \\ framework \cite{CycleGAN2017,isola2017image} supports efficient training, fine-tuning, and inference for Pix2Pix. This PyTorch-based tool provides pre-trained models, GPU acceleration, and flexible configurations, enabling industry professionals to quickly adapt the framework for diverse image-to-image translation tasks.

\subsubsection{CycleGAN: Unpaired Image-to-Image Translation}

CycleGAN \cite{CycleGAN2017} is a GAN-based framework specifically designed for unpaired image-to-image translation tasks, where there is no direct correspondence between input and target images. Unlike paired methods such as Pix2Pix, CycleGAN learns to translate between two domains using unaligned datasets. This flexibility makes it particularly valuable in scenarios where obtaining paired data is challenging or expensive. CycleGAN has been widely adopted for applications such as artistic style transfer, seasonal image transformation, and domain adaptation in computer vision.

\begin{figure}[h]
    \centering
    \includegraphics[width=1\textwidth]{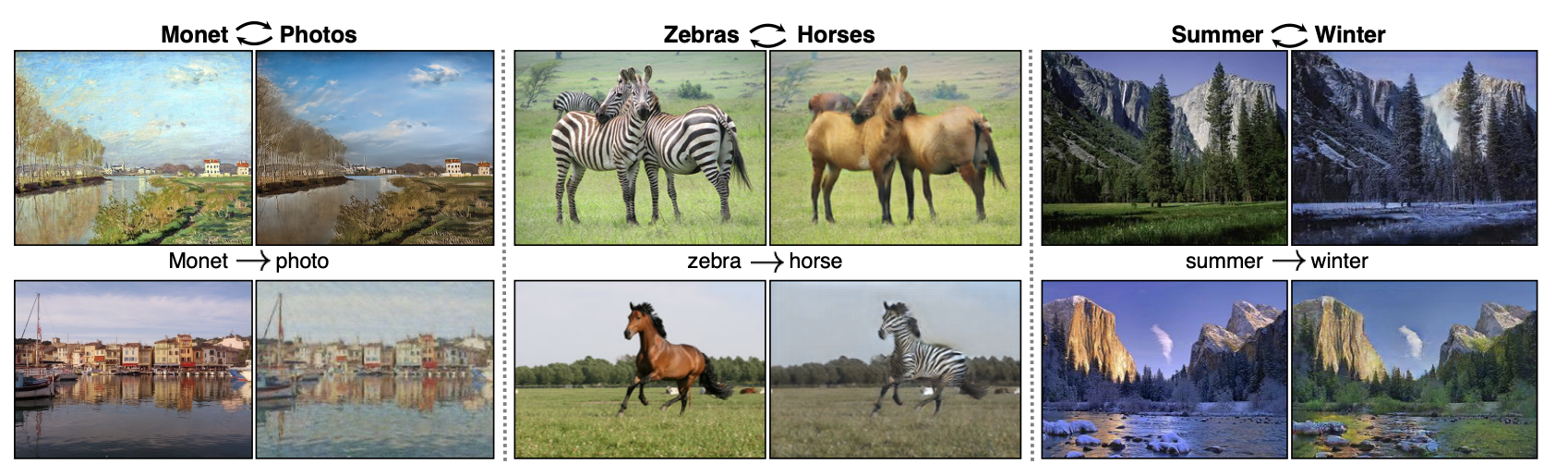}
    \caption{CycleGAN enables unpaired image-to-image translation tasks, such as transforming horse images to zebras, and vice versa, or changing the style of landscapes between summer and winter \cite{CycleGAN2017}.}
    \label{fig:cyclegan_translation}
\end{figure}

\paragraph{Key Features of CycleGAN}
\begin{itemize}
    \item \textbf{Unpaired Training:} Eliminates the need for paired datasets, making it suitable for applications where aligned data is unavailable.
    \item \textbf{Cycle-Consistency Loss:} Ensures that translating an image to another domain and back preserves the original content and structure.
    \item \textbf{Bidirectional Translation:} Uses two generators and two discriminators to perform translations in both directions between two domains.
    \item \textbf{Versatility:} Applicable to a wide range of tasks, including style transfer, domain adaptation, and image enhancement.
\end{itemize}

\begin{figure}[htbp]
    \centering
    \includegraphics[width=1\textwidth]{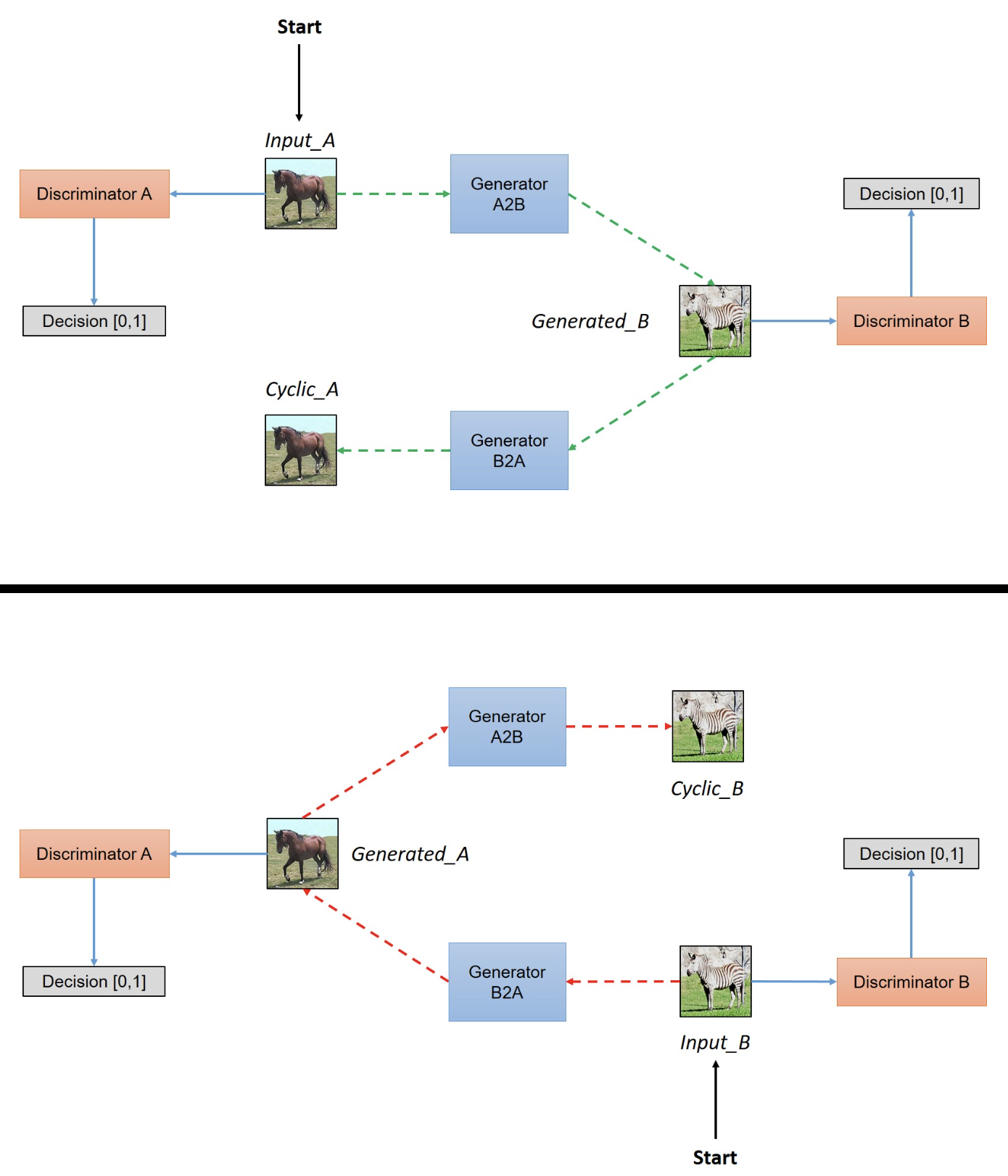}
    \caption{Simplified architecture of CycleGAN. Generators ($G_{XY}$, $G_{YX}$) perform bidirectional translation, while discriminators ($D_X$, $D_Y$) evaluate the realism of generated images \cite{CycleGAN2017}.}
    \label{fig:cyclegan_architecture}
\end{figure}

\paragraph{How CycleGAN Works (Simplified):} CycleGAN consists of two generator networks ($G_{XY}$ and $G_{YX}$) and two discriminator networks ($D_X$ and $D_Y$). The generators perform translations between domains $X$ and $Y$, while the discriminators evaluate the realism of the generated images in their respective domains. A cycle-consistency loss ensures that translating an image to the other domain and back retains the original content. This dual loss mechanism combines adversarial loss (for realism) with cycle-consistency loss (for content preservation), enabling effective and reliable domain translation. Figure~\ref{fig:cyclegan_architecture} illustrates the simplified architecture of CycleGAN.

\paragraph{Practical Applications of CycleGAN}
\begin{enumerate}
    \item \textbf{Artistic Style Transfer}
    \begin{itemize}
        \item \textbf{Use Case:} Transforming photographs into paintings in the style of artists like Monet or Van Gogh.
        \item \textbf{Process:} CycleGAN maps unpaired datasets of photos and artworks, transferring styles while preserving content.
        \item \textbf{Impact:} Enables quick creation of stylized content for art, design, and media.
    \end{itemize}

    \item \textbf{Self-Driving Cars: Domain Adaptation}
    \begin{itemize}
        \item \textbf{Use Case:} Translating synthetic road scenes into realistic images for training models.
        \item \textbf{Process:} CycleGAN converts synthetic visuals into realistic data, bridging the gap between simulations and real-world conditions.
        \item \textbf{Impact:} Enhances training data, improving self-driving car performance and safety.
    \end{itemize}

    \item \textbf{Healthcare: Cross-Modality Image Translation}
    \begin{itemize}
        \item \textbf{Use Case:} Converting MRI scans into CT scans for multimodal analysis.
        \item \textbf{Process:} CycleGAN learns unpaired translations between imaging modalities while retaining diagnostic details.
        \item \textbf{Impact:} Simplifies access to complementary imaging insights, improving diagnostics.
    \end{itemize}
\end{enumerate}

\paragraph{Implementation Framework:} Similar to Pix2Pix,  the following the framework \texttt{pytorch-CycleGAN-and-pix2pix}  \cite{CycleGAN2017} can be leveraged for CycleGAN training, fine-tuning, and inference. This PyTorch-based tool provides pre-trained models, GPU acceleration, and flexible configurations, enabling efficient application of CycleGAN to a variety of image-to-image translation tasks.

\subsubsection{StyleGAN: High-Quality Image Synthesis}

StyleGAN \cite{karras2019style} is a GAN-based framework designed for generating high-quality, high-resolution images with an emphasis on style control and disentangled representations. Unlike other GAN-based models, StyleGAN focuses on synthesizing entirely new content within a domain while providing unparalleled control over image attributes, such as color, texture, and structure. This capability makes it a valuable tool for applications requiring creative and detailed image generation.

While Pix2Pix and CycleGAN address image-to-image translation tasks, StyleGAN’s purpose and architecture diverge. It is not designed for transforming existing images between domains but for generating novel images from latent representations. Its unique style-based architecture enables fine-grained manipulation of specific visual features, offering flexibility and precision unmatched by translation-focused GANs.

\paragraph{Key Features of StyleGAN}
\begin{itemize}
    \item \textbf{Style-Based Architecture:} StyleGAN introduces a novel architecture where latent variables are transformed into "style" vectors, influencing different layers of the generator to control specific image features.
    \item \textbf{Progressive Growing:} Initially proposed in the predecessor model, this technique trains the network to generate low-resolution images and progressively increases resolution, improving stability and quality.
    \item \textbf{Disentangled Representations:} The model’s architecture enables fine-grained control over attributes, such as facial expressions, hairstyle, or lighting in generated images.
    \item \textbf{Adaptive Instance Normalization (AdaIN):} A key innovation where style vectors modulate features at each layer of the generator, enabling smooth and intuitive style manipulation.
\end{itemize}

\paragraph{How StyleGAN Works (Simplified):}StyleGAN generates high-quality images by training on a dataset of real images. Random noise vectors are transformed into latent codes, which are processed into style vectors controlling image attributes like texture and structure. These style vectors are injected into the generator using Adaptive Instance Normalization (AdaIN), modifying features at different scales. A discriminator evaluates the realism of generated images, refining the generator through adversarial training. This process enables precise control over visual attributes, producing realistic and customizable images.

\begin{figure}[h]
    \centering
    \includegraphics[width=1\textwidth]{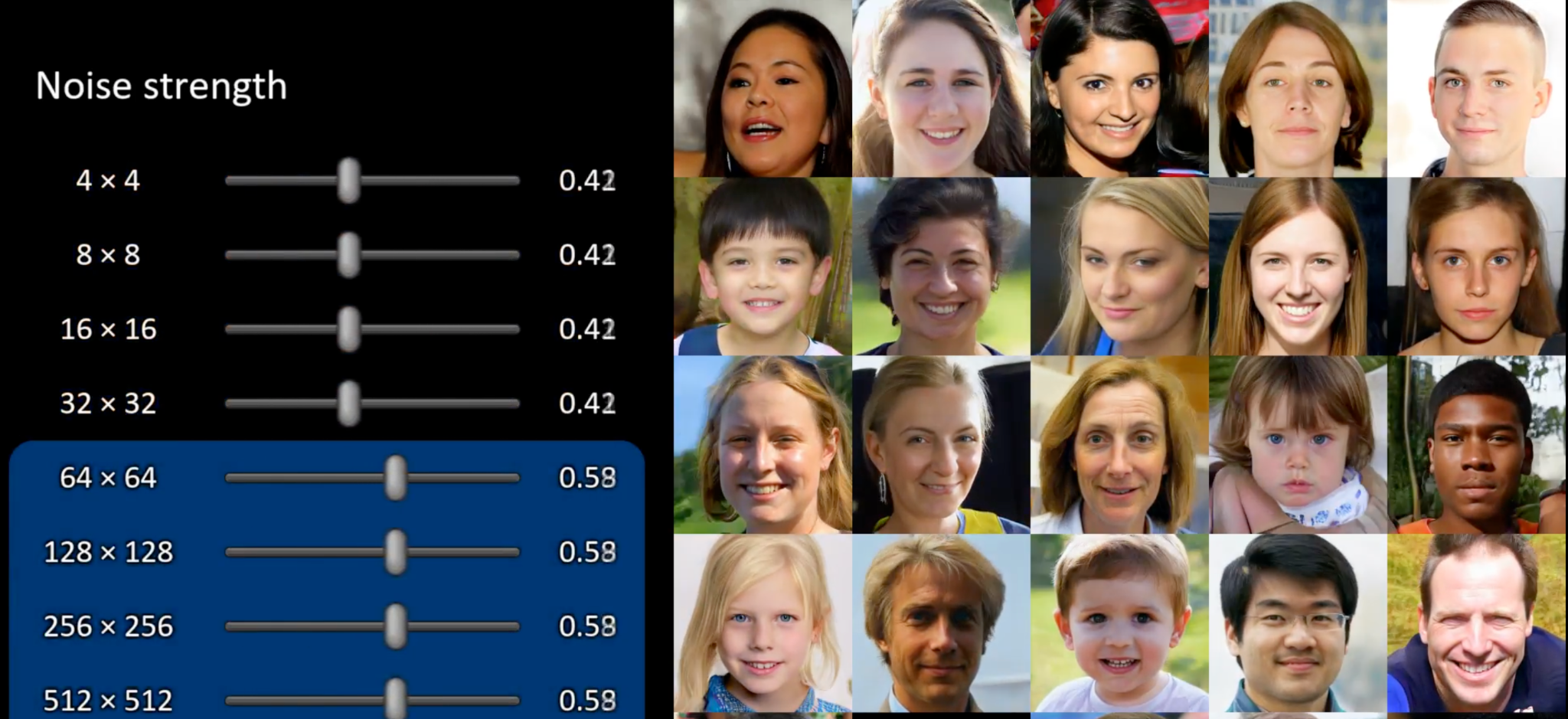} 
    \caption{Example outputs from StyleGAN showing synthetic faces generated at various resolutions. These images demonstrate the model’s ability to produce highly realistic visuals from low to high resolutions, while maintaining control over style and noise strength \cite{karras2019style}.}
    \label{fig:stylegan_outputs}
\end{figure}

\paragraph{Applications of StyleGAN}
\begin{enumerate}
    \item \textbf{E-Commerce and Augmented Shopping}
    \begin{itemize}
        \item \textbf{Use Case:} Generating virtual try-ons for clothing or accessories.
        \item \textbf{Process:} StyleGAN is trained on product images to generate realistic, diverse representations of items on various body types or in different settings, providing a virtual try-on experience.
        \item \textbf{Impact:} Enhances online shopping by offering personalized previews, reducing return rates, and increasing customer satisfaction.
    \end{itemize}

    \item \textbf{Application Filters for Visual Effects}
    \begin{itemize}
        \item \textbf{Use Case:} Creating advanced photo or video filters for social media apps.
        \item \textbf{Process:} StyleGAN generates realistic transformations, such as background alterations or artistic effects, using its ability to manipulate visual attributes with precision.
        \item \textbf{Impact:} Provides high-quality, creative effects, boosting user engagement and content creation.
    \end{itemize}

    \item \textbf{Supply Chain and Product Inspection}
    \begin{itemize}
        \item \textbf{Use Case:} Generating synthetic product images with defects for training inspection systems.
        \item \textbf{Process:} StyleGAN is fine-tuned on datasets of product images to synthesize realistic defect scenarios, such as scratches or missing components, augmenting data for defect detection models.
        \item \textbf{Impact:} Improves quality inspection accuracy, reducing operational costs and product recalls.
    \end{itemize}
\end{enumerate}

StyleGAN represents a significant advancement in generative modeling by combining high-quality image synthesis with controllable style manipulation. Its applications span multiple industries, from media and entertainment to research, education, e-commerce, and supply chain management. For more technical details, readers can reference the original paper \cite{karras2019style}.

\subsection{Diffusion Models: A Modern Paradigm in Generative Modeling}
\label{sec:diffusion_models}

Diffusion models \cite{ho2020denoising,yang2024diffusionmodelscomprehensivesurvey} represent a powerful class of generative models that produce high-quality data by modeling the denoising process in reverse. Unlike GANs, which employ adversarial training between a generator and discriminator, diffusion models use probabilistic processes to gradually transform noise into structured outputs, making them highly stable during training and capable of generating diverse and realistic samples.

\paragraph{Key Differences with GANs:} Diffusion models differ fundamentally from GANs in their training approach and generative process. While GANs rely on adversarial dynamics, diffusion models avoid such instability by employing probabilistic methods based on noise and denoising. They generate data by iteratively refining noise into structured outputs, whereas GANs map noise to data in a single step. This iterative nature also enables diffusion models to produce more diverse samples, making them particularly effective in applications requiring fine detail and variety.

\paragraph{Key Features of Diffusion Models: }
\begin{itemize}
    \item \textbf{Probabilistic Framework:} They rely on Markov chains to simulate a gradual transformation between noise and data distributions.
    \item \textbf{Denoising Process:} A reverse process iteratively removes noise, reconstructing data with high fidelity.
    \item \textbf{Stability and Diversity:} Diffusion models minimize well-defined objectives, resulting in stable training and diverse output.
    \item \textbf{Scalability:} They can handle large datasets and generate high-resolution outputs.
\end{itemize}

\begin{figure}[htbp]
    \centering
    \includegraphics[width=1\textwidth]{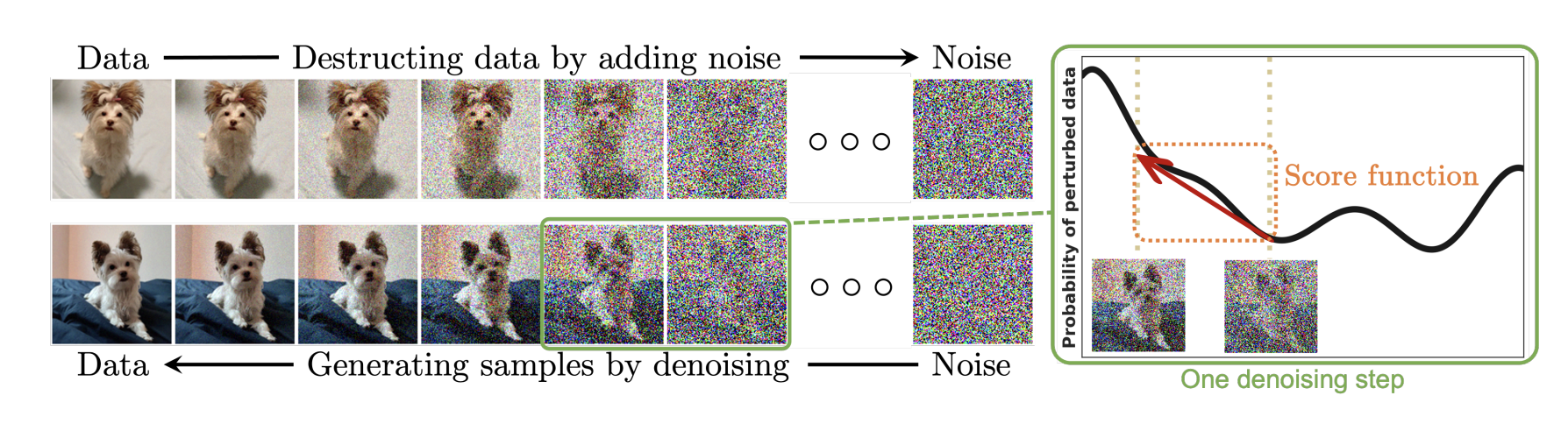}
    \caption{Diffusion models operate by progressively adding noise to the data (forward process) and then learning to reverse this process (reverse denoising). The reverse process reconstructs the data step-by-step by estimating the score function, which guides the model toward regions of higher likelihood \cite{yang2024diffusionmodelscomprehensivesurvey}.}
    \label{fig:diffusion_process}
\end{figure}

\paragraph{How It Works (Simplified):} Diffusion models operate in two main phases: a forward process and a reverse process. In the forward process, data samples are progressively corrupted with noise through a sequence of steps, effectively transforming them into random noise. The reverse process then learns to predict and remove the noise in incremental steps, reconstructing the original data. The training objective involves minimizing the difference between the model's predicted noise and the actual noise added during the forward process. By iteratively refining noisy inputs, diffusion models excel at generating detailed and realistic outputs. Figure \ref{fig:diffusion_process} illustrates how diffusion models add noise in the forward process and remove it step-by-step in the reverse, guided by the score function to reconstruct realistic outputs \cite{yang2024diffusionmodelscomprehensivesurvey}.

\paragraph{Types of Diffusion Models:} 
Diffusion models have evolved into several key types, each tailored for specific applications. Denoising Diffusion Probabilistic Models (DDPMs), introduced by Ho et al. \cite{ho2020denoising}, are foundational and model the reverse denoising process step by step, offering stability and high-quality outputs. 

Score-based models \cite{song2020score} build upon this framework by learning the gradient of the data distribution, enabling more flexible sampling techniques such as Langevin dynamics and stochastic differential equations (SDEs). These models are particularly effective in tasks requiring continuous-time formulations.

Latent Diffusion Models (LDMs) \cite{rombach2022high} compress the diffusion process into a latent space, significantly reducing computational requirements while maintaining high-resolution output quality. This innovation makes LDMs well-suited for applications such as text-to-image generation and high-resolution video synthesis.

Variance Exploding and Variance Preserving Diffusion Models \cite{song2020score} provide alternate noise scheduling strategies in the forward process, enabling better trade-offs between training stability and sample diversity. These variations allow finer control over the diffusion dynamics.

Conditional Diffusion Models \cite{saharia2022imagen} extend the basic framework by incorporating additional information, such as text, images, or other modalities, into the generation process. These models have been pivotal in applications like text-to-image synthesis and image editing.

Guided Diffusion Models \cite{dhariwal2021diffusion} use external guidance, such as classifiers or other pre-trained models, to steer the generative process toward desired attributes or outcomes. This approach enables fine-grained control over the generated samples and improves alignment with specific user-defined prompts.

More recently, Consistency Models \cite{song2023consistency} aim to accelerate the sampling process by directly learning a consistency function that avoids the need for iterative reverse diffusion steps. These models significantly reduce computational cost while maintaining high-quality outputs.

With these diverse approaches, diffusion models continue to expand their applicability across domains, including creative content generation, scientific visualization, and industrial quality inspection.

\paragraph{Practical Applications}
\begin{enumerate}
    \item \textbf{Manufacturing: Defect Detection in Noisy Environments}
    \begin{itemize}
        \item \textbf{Use Case:} Enhancing the quality of noisy inspection images to identify defects in products during manufacturing.
        \item \textbf{Process:} Diffusion models denoise inspection images by iteratively removing noise while preserving fine details, enabling accurate detection of scratches, cracks, or other anomalies.
        \item \textbf{Impact:} Improves the accuracy of defect detection systems, reduces false positives, and ensures consistent product quality in high-throughput manufacturing.
    \end{itemize}

    \item \textbf{Healthcare: Medical Imaging Enhancement}
    \begin{itemize}
        \item \textbf{Use Case:} Restoring clarity in noisy or low-quality medical images such as MRI or CT scans, enabling better diagnostics.
        \item \textbf{Process:} Diffusion models are trained to reconstruct high-quality medical images from noisy inputs, refining subtle details crucial for medical interpretation.
        \item \textbf{Impact:} Enhances diagnostic accuracy, reduces repeat scans, and supports cost-effective healthcare delivery.
    \end{itemize}

    \item \textbf{Surveillance: Low-Light Image Enhancement}
    \begin{itemize}
        \item \textbf{Use Case:} Improving visibility in surveillance footage captured in low-light or noisy conditions.
        \item \textbf{Process:} Diffusion models are applied to enhance noisy, low-light images by recovering missing details and refining textures through iterative denoising.
        \item \textbf{Impact:} Enables reliable monitoring and analysis in security systems, even in challenging environments, ensuring critical details are not lost.
    \end{itemize}
\end{enumerate}

Diffusion models have emerged as a robust alternative to GANs, excelling in tasks that demand stable training and diverse output generation. Their applications span creative industries, healthcare, and multimedia, establishing them as a cornerstone of modern generative modeling.

\section{Latent Space Representations: Variational Autoencoders}  

\label{VAEs}
Variational Autoencoders (VAEs) \cite{kingma2013autoencoding} are a foundational class of generative models that use latent space representations to encode and decode data through probabilistic frameworks. By combining deep neural networks with Bayesian inference, VAEs enable smooth, continuous latent spaces that facilitate interpolation, attribute manipulation, and structured generation. Unlike deterministic autoencoders, VAEs regularize the latent space to conform to a predefined probability distribution (e.g., Gaussian), ensuring meaningful sampling and robust generalization.

\subsection{Key Features of VAEs}  
\begin{itemize}
    \item \textbf{Probabilistic Latent Variables:} VAEs map input data to latent variables drawn from a probabilistic distribution, typically a multivariate Gaussian, allowing for smooth interpolation and meaningful sampling.
    \item \textbf{Encoder-Decoder Framework:} The encoder compresses input data into latent representations, while the decoder reconstructs the original data from these latent variables.
    \item \textbf{KL Regularization:} A Kullback-Leibler (KL) divergence term ensures the latent variables conform to a prior distribution, promoting disentanglement and preventing overfitting.
    \item \textbf{Interpretable Latent Space:} The structured latent space enables fine-grained manipulation of attributes, such as object styles, facial expressions, or colors.
\end{itemize}

\subsection{How VAEs Work?}  
VAEs employ a probabilistic encoder-decoder framework. The encoder compresses input data into a distribution over a latent space, represented by mean and variance vectors. Latent variables are sampled from this distribution and fed into the decoder, which reconstructs the original data. The objective function comprises two components: (1) the reconstruction loss, measuring how accurately the generated data matches the input, and (2) the KL divergence loss, ensuring that the latent space aligns with a prior distribution, typically a standard Gaussian. This dual objective encourages both fidelity and structured representations.

The VAE objective function can be expressed as follows:
\begin{equation}
    \mathcal{L}_{\text{VAE}}(\theta, \phi) = \mathbb{E}_{q_\phi(z|x)} \left[ \log p_\theta(x|z) \right] - \text{KL} \left( q_\phi(z|x) \, || \, p(z) \right)
\end{equation}

\paragraph{Explanation of Parameters:}
\begin{itemize}
    \item $x$: Input data, which is encoded into the latent space and reconstructed by the decoder.
    \item $z$: Latent variable sampled from the posterior distribution $q_\phi(z|x)$.
    \item $\mathcal{L}_{\text{VAE}}$: The total loss function for training the VAE.
    \item $p_\theta(x|z)$: The likelihood of reconstructing $x$ given the latent variable $z$, modeled by the decoder. This term corresponds to the reconstruction loss.
    \item $q_\phi(z|x)$: The posterior distribution of the latent variable $z$, modeled by the encoder.
    \item $p(z)$: The prior distribution over the latent space, typically a standard Gaussian $\mathcal{N}(0, I)$.
    \item $\text{KL} \left( q_\phi(z|x) \, || \, p(z) \right)$: The Kullback-Leibler (KL) divergence, which measures how closely the posterior $q_\phi(z|x)$ matches the prior $p(z)$.
\end{itemize}

The first term in the objective function, $\mathbb{E}_{q_\phi(z|x)} \left[ \log p_\theta(x|z) \right]$, ensures that the decoder reconstructs data accurately, penalizing any deviation between the input $x$ and its reconstruction. The second term, $\text{KL} \left( q_\phi(z|x) \, || \, p(z) \right)$, acts as a regularizer, aligning the learned latent distribution with the prior to ensure a smooth and interpretable latent space. Together, these terms balance data fidelity and latent space structure, allowing the VAE to generate meaningful and diverse outputs.

\subsection{Types of Variational Autoencoders}
Over time, several variants of VAEs have been developed to address specific challenges and expand their applications:
\begin{itemize}
    \item \textbf{Beta-VAEs:} Introduced by Higgins et al. \cite{higgins2017beta}, Beta-VAEs modify the standard VAE framework by introducing a weighting factor ($\beta$) for the KL divergence term. This encourages greater disentanglement in the latent space, allowing for better control over independent attributes of the data.
    \item \textbf{Conditional VAEs (CVAEs):} CVAEs extend the VAE framework by incorporating conditional inputs, such as labels or auxiliary data, to guide the generation process. This has proven effective in tasks like image-to-image translation \cite{sohn2015learning}.
\item \textbf{Vector-Quantized VAEs (VQ-VAEs):} Proposed by Oord et al. \cite{oord2017vqvae}, VQ-VAEs replace the continuous latent space with a discrete codebook of latent embeddings, improving the quality of generated outputs in tasks such as high-resolution image synthesis and audio generation. Figure \ref{fig:vqvae} illustrates the mechanism of VQ-VAEs, where the encoder maps input data to the nearest discrete embedding in the codebook, while the gradients adjust the configuration for subsequent iterations.

\begin{figure}[htbp]
    \centering
    \includegraphics[width=1.1\textwidth]{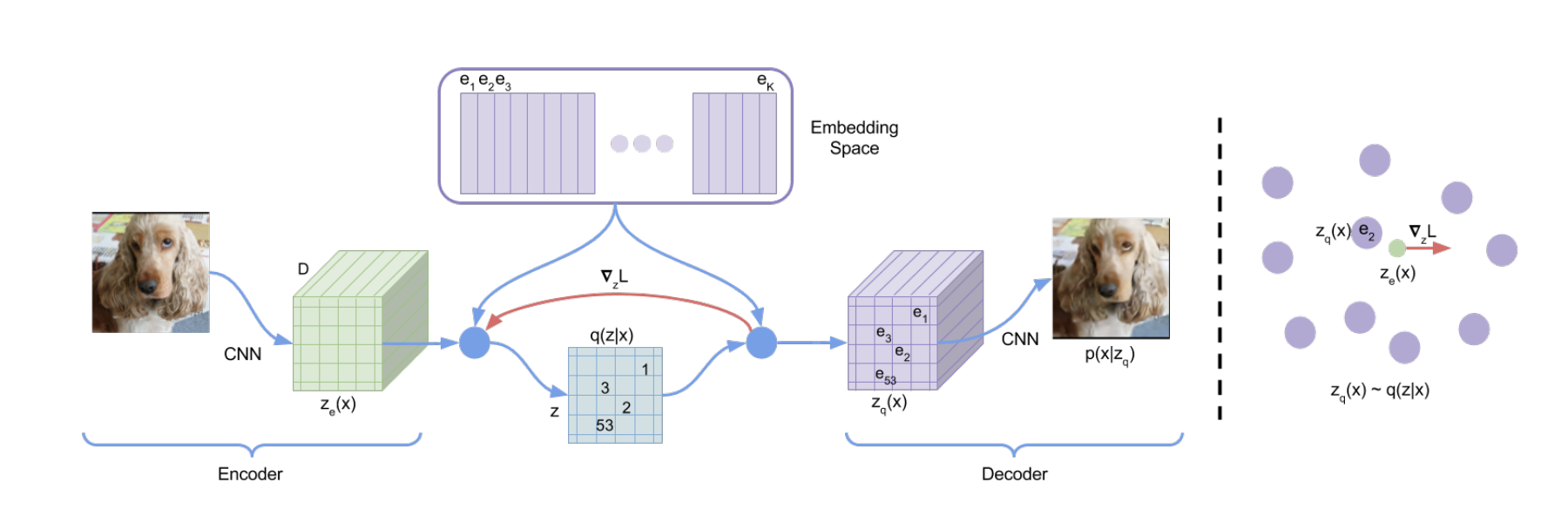}
    \caption{A figure describing the VQ-VAE. Left: The encoder output $z(x)$ is mapped to the nearest point $e_2$ in the discrete codebook. Right: A visualization of the embedding space, where the gradient $\nabla_z L$ (in red) pushes the encoder to change its output, potentially altering the configuration in the next forward pass \cite{oord2017vqvae}.}
    \label{fig:vqvae}
\end{figure}

    \item \textbf{Hierarchical VAEs:} Hierarchical VAEs \cite{sonderby2016ladder} introduce multiple levels of latent variables, capturing both coarse and fine-grained details. This approach enhances scalability and enables high-resolution data generation.
    \item \textbf{Denoising VAEs:} By introducing noise during the encoding process, denoising VAEs improve robustness and handle noisy or incomplete data effectively \cite{im2017denoising}.
    \item \textbf{Sparse VAEs:} Sparse VAEs \cite{makhzani2015adversarial} enforce sparsity in the latent space, promoting interpretability and reducing redundancy in latent variables.
\end{itemize}

\subsection{Applications of VAEs}  
\begin{enumerate}

    \item \textbf{Medical Imaging Enhancement}
    \begin{itemize}
        \item \textbf{Use Case:} Improving diagnostic accuracy by reconstructing high-quality medical images and detecting anomalies in radiology and pathology.
        \item \textbf{Process:} VAEs are trained on medical datasets to learn latent representations, enabling the generation of clean reconstructions, anomaly detection, and synthetic datasets for training AI models.
        \item \textbf{Impact:} Enhances diagnostic precision, supports clinical decision-making, and reduces reliance on large labeled datasets.
    \end{itemize}

    \item \textbf{Image Interpolation and Morphing}
    \begin{itemize}
        \item \textbf{Use Case:} Creating seamless transitions between images for applications in visual effects, design, and multimedia.
        \item \textbf{Process:} VAEs interpolate between latent space embeddings of input images, generating smooth transformations that blend attributes from both inputs.
        \item \textbf{Impact:} Enables novel visual effects and morphing capabilities, widely used in entertainment, advertising, and design.
    \end{itemize}

    \item \textbf{Representation Learning for Vision Tasks}
    \begin{itemize}
        \item \textbf{Use Case:} Learning compact and meaningful representations for tasks like clustering, anomaly detection, and dimensionality reduction.
        \item \textbf{Process:} VAEs encode high-dimensional data into low-dimensional latent representations, which are used for downstream tasks like segmentation or classification.
        \item \textbf{Impact:} Simplifies complex vision tasks, enhances interpretability, and improves model performance in unsupervised or semi-supervised settings.
    \end{itemize}

\end{enumerate}

Variational Autoencoders provide a robust framework for generative modeling, combining flexibility, interpretability, and scalability. Continued innovations, such as hierarchical and discrete latent spaces, are expected to further expand their utility across diverse domains.

\newpage

\section{Prompt-to-Image Generation}
\label{prompt}

Prompt-to-image generation is a subcategory of generative modeling that transforms natural language descriptions into realistic and semantically coherent images. Unlike noisy vector-based image generation \ref{Noisy}, which uses random noise as input (e.g., GANs), or image-to-image translation tasks, which conditionally transform one image into another (e.g., CycleGAN), prompt-to-image generation bridges linguistic and visual modalities. This unique capability makes it a cornerstone of multimodal AI, redefining content creation across industries.

Recent advancements in this field, including models like DALL-E\cite{ramesh2022openai}, Imagen\cite{saharia2022imagen}, DeepSeek Janus-Pro \cite{chen2025januspro}, and Stable Diffusion \cite{rombach2022high}, have demonstrated its potential to revolutionize creative workflows. These tools enable designers, artists, and professionals to prototype concepts rapidly and generate tailored visuals from simple textual descriptions. This paradigm has democratized access to high-quality, customizable visual content, fostering innovation in design, multimedia, and entertainment.

\begin{figure}[htbp]
    \centering
    \includegraphics[width=\textwidth]{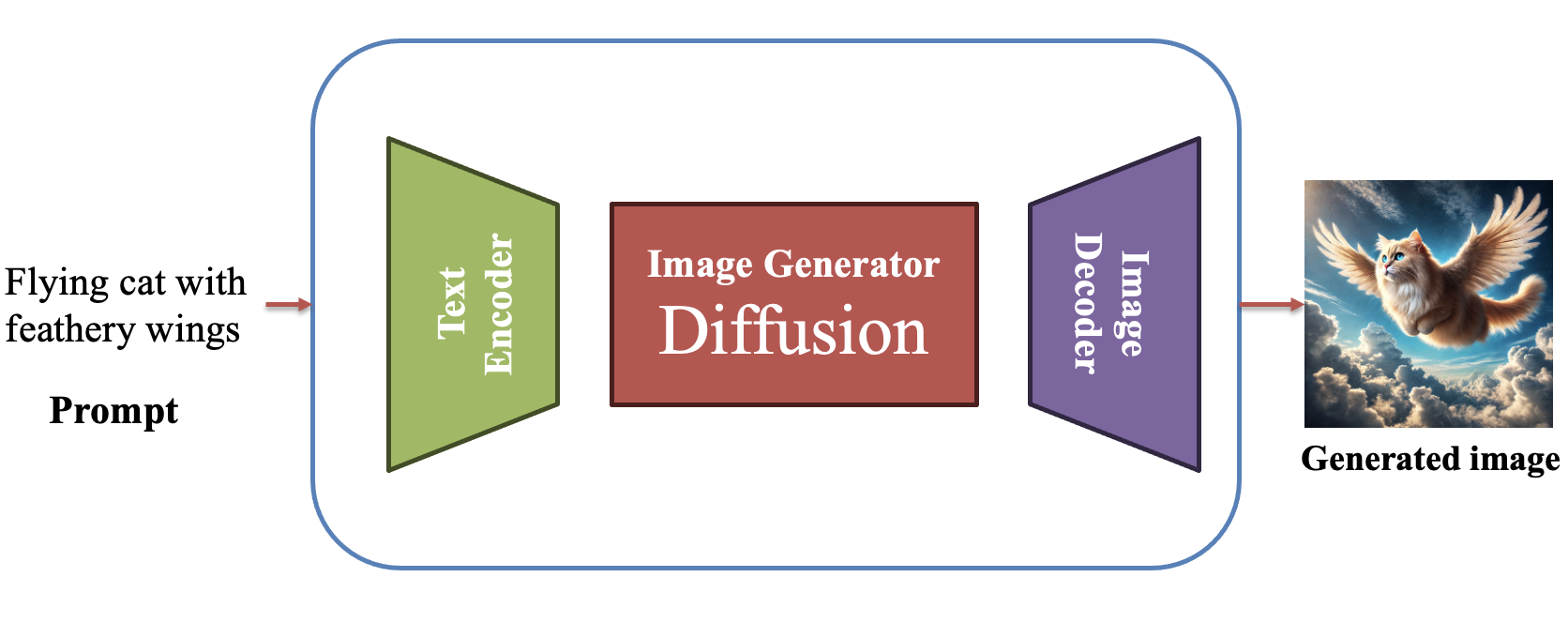}
    \caption{Core components of the prompt-to-image generation framework. The text encoder processes the natural language input into semantic embeddings. The image generator translates these embeddings into a latent representation, which is refined and decoded into a high-quality image.}
    \label{fig:core_components}
\end{figure}

The process of prompt-to-image generation relies on three core components, as illustrated in Figure~\ref{fig:core_components}: the \textbf{text encoder}, the \textbf{image generator}, and the \textbf{image decoder}. The text encoder translates the natural language prompt into a semantic representation, which is aligned with a latent image space by the image generator. The image decoder then transforms the latent representation into a full-resolution, high-quality visual output. Together, these components create a seamless pipeline for converting language into imagery, enabling precise and creative control over the generated content.

In the following sections, we will explore each of these components in detail, discussing their roles, mechanisms, and implementations in state-of-the-art models such as Stable Diffusion,  OpenAI's DALL-E and  DeepSeek Janus-Pro.

\subsection{Text Encoder}

The text encoder is a fundamental component in prompt-to-image generation frameworks, responsible for transforming natural language prompts into high-dimensional semantic representations. This transformation allows the downstream components, such as the image generator and decoder, to interpret the textual input effectively. By capturing the nuances, context, and intent of a prompt, the text encoder ensures that the generated image aligns with the user's requirements.

Text encoders can be broadly categorized into two types: \textbf{pretrained text encoders}, which are trained on textual corpora and capture linguistic semantics, and \textbf{multimodal encoders}, which are specifically designed to align text and image representations into a shared embedding space.

\subsubsection{Pretrained Text Encoders: BERT and Family}

Pretrained text encoders are models trained on large-scale textual datasets to capture the syntactic and semantic properties of language. These models produce embeddings that encode the meaning and context of input text, enabling their use in various natural language processing (NLP) and generation tasks.

\begin{itemize}
    \item \textbf{BERT (Bidirectional Encoder Representations from Transformers):}
    Introduced by Devlin et al. \cite{devlin2018bert}, BERT is a transformer-based encoder that learns bidirectional contextual embeddings by leveraging a masked language modeling (MLM) objective. BERT's ability to capture context from both directions makes it highly effective for understanding complex prompts in tasks such as prompt-to-image generation. Variants of BERT, such as RoBERTa \cite{liu2019roberta} and ALBERT \cite{lan2019albert}, further optimize performance through architectural and training improvements.

    \item \textbf{GPT (Generative Pretrained Transformer):}
    Although GPT \cite{radford2019gpt2} is primarily known for its autoregressive generation capabilities, its encoder-like component can also be leveraged for prompt embedding. However, unlike BERT, GPT captures unidirectional context, which may limit its ability to encode complex interrelations in textual prompts effectively.
\end{itemize}

While pretrained text encoders like BERT excel at capturing linguistic semantics, they do not inherently align textual representations with visual concepts. This limitation can be addressed by using multimodal encoders, which are explicitly trained for cross-modal tasks.

\subsubsection{Multimodal Encoders}
\label{Multimodal_encoder}
Multimodal encoders are designed to align textual and visual embeddings within a unified space, enabling superior performance in tasks that require a connection between language and images. These models are pretrained on large-scale datasets containing paired image-text examples, allowing them to understand and map semantic similarities across modalities.

\begin{itemize}
    \item \textbf{CLIP: Contrastive Language-Image Pretraining):}
CLIP (Contrastive Language–Image Pretraining) \cite{radford2021clip} aligns text and image embeddings into a shared semantic space through contrastive learning. Using two separate encoders—a text encoder and an image encoder—it processes text and images into embedding vectors and learns to maximize similarity for matching pairs while minimizing it for non-matching pairs. This alignment allows the shared space to capture visual semantics, making CLIP effective for tasks such as text-to-image generation. As illustrated in Figure~\ref{fig:clip_alignment} \cite{radford2021clip}, the text encoder transforms text (e.g., "Pepper the Aussie pup") into embedding vectors ($T_1, T_2, \dots$), while the image encoder maps images into corresponding embeddings ($I_1, I_2, \dots$). A similarity matrix is computed, ensuring that matching text-image pairs (e.g., the text and an image of the same dog) are closer in the embedding space, while non-matching pairs are pushed apart.

\begin{figure}[htbp]
    \centering
    \includegraphics[width=1.05\textwidth]{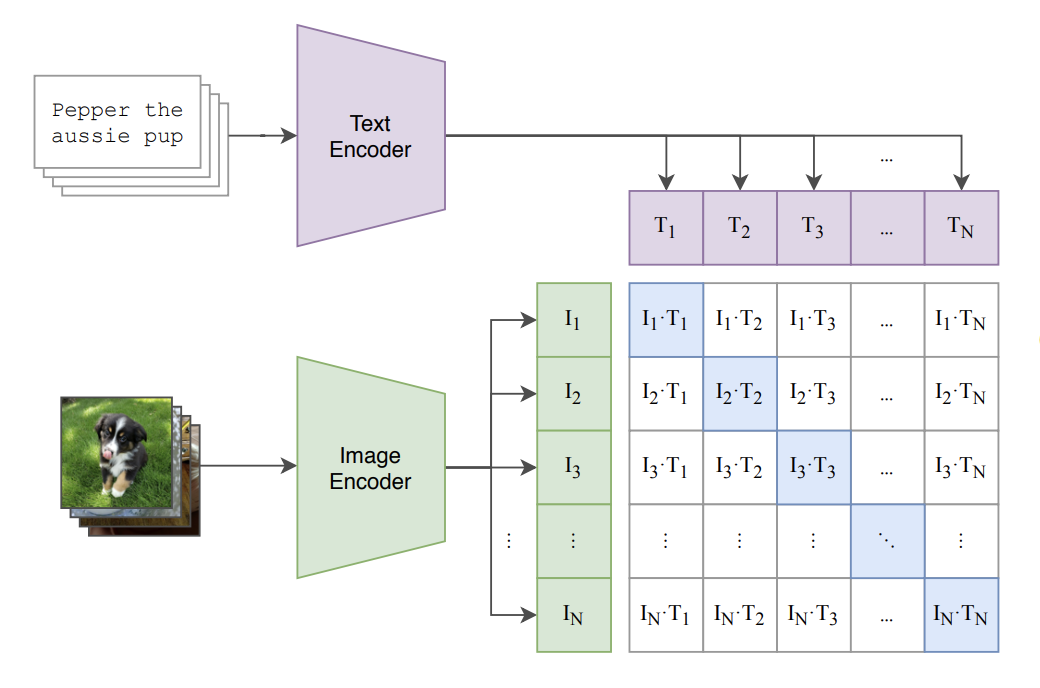}
    \caption{Contrastive pretraining in CLIP: Text and image embeddings are aligned into a shared space where visual semantics are preserved. Matching pairs (e.g., "Pepper the Aussie pup" and its image) have higher similarity scores \cite{radford2021clip}.}
    \label{fig:clip_alignment}
\end{figure}

This alignment mechanism ensures semantic consistency between text and images, enabling models like CLIP to serve as powerful foundations for prompt-to-image generation tasks.

    \item \textbf{FLAVA:}
    FLAVA \cite{singh2022flava} extends the capabilities of multimodal encoders by combining vision, language, and vision-language objectives. It is trained on diverse datasets and supports various cross-modal tasks, making it a versatile encoder for prompt-to-image applications.

    \item \textbf{Flamingo:}
    Flamingo \cite{alayrac2022flamingo} is a multimodal encoder-decoder model that integrates visual and textual inputs in a unified space. Unlike CLIP, Flamingo supports more complex interactions between modalities, such as generating captions from images or synthesizing images from text prompts.

    \item \textbf{CoCa:}
Contrastive Captioner (CoCa) \cite{yu2022coca} is a unified vision-language model that combines contrastive learning and generative pretraining. CoCa jointly learns image-text alignment and caption generation by training on paired image-text datasets. This dual-objective approach enables CoCa to perform well on a wide range of cross-modal tasks, including text-to-image generation, image captioning, and retrieval. CoCa’s integration of both discriminative and generative objectives makes it a highly adaptable and powerful encoder for multimodal applications.

    \item \textbf{Other Multimodal Encoders:}
    Models like ALIGN \cite{jia2021scaling} and BLIP \cite{li2022blip} further push the boundaries of multimodal alignment by scaling dataset size and introducing innovative architectural designs. These encoders are particularly effective for cross-modal tasks requiring precise semantic mapping between language and images.

\end{itemize}

Multimodal encoders are inherently superior for prompt-to-image generation tasks compared to pretrained text encoders. Their ability to align text and images into a shared embedding space ensures that the generated images are semantically coherent and contextually relevant to the input prompts. These models represent the cutting edge of text encoding for multimodal generative frameworks.

\subsection{Image Generator}

The image generator lies at the heart of the prompt-to-image generation framework, translating semantic representations from the text encoder into latent image embeddings. Advanced generative architectures, such as diffusion models, dominate this component due to their ability to produce high-quality, realistic outputs. Diffusion models, as discussed in depth in Section~\ref{sec:diffusion_models}, iteratively refine noise into meaningful latent representations using probabilistic denoising techniques. Their flexibility and scalability make them well-suited for generating diverse and complex visual patterns.

In the context of prompt-to-image generation, the image generator integrates seamlessly with the text encoder by conditioning the diffusion process on the semantic embeddings of the input prompt. This ensures that the generated latent representation reflects the text's intent and context, paving the way for accurate and coherent image synthesis.

\subsection{Image Decoder}

The image decoder serves as the final stage of the prompt-to-image generation pipeline, transforming the latent image embeddings produced by the generator into high-resolution visual outputs. As detailed in Section~\ref{VAEs}, the decoder component often leverages techniques from variational autoencoders (VAEs) to reconstruct pixel-level details from latent spaces.

In prompt-to-image frameworks, the decoder is tightly integrated with the generator, ensuring that the refined latent embeddings are translated into visually coherent and semantically aligned images. The decoding process is crucial for maintaining image quality and resolution, making it indispensable in achieving photorealistic or stylistically unique outputs that adhere to the user's prompt.

By leveraging the latent-to-image reconstruction capabilities of decoders, prompt-to-image generation frameworks bridge the gap between abstract semantic representations and high-quality, tangible visuals. This integration underscores the critical role of a robust decoding mechanism in ensuring the fidelity and resolution of the final output within the overall framework.

\subsection{Stable Diffusion: Diffusion-Based Frameworks}

Stable Diffusion \cite{rombach2022high} is a diffusion-based generative framework designed to produce high-resolution, diverse, and semantically coherent images. Unlike traditional diffusion models that operate directly in pixel space, Stable Diffusion performs the diffusion process in a compressed latent space, significantly reducing computational costs while preserving image quality.

\paragraph{How Stable Diffusion Works:}

\begin{figure}[htbp]
    \centering
    
    \includegraphics[width=\textwidth]{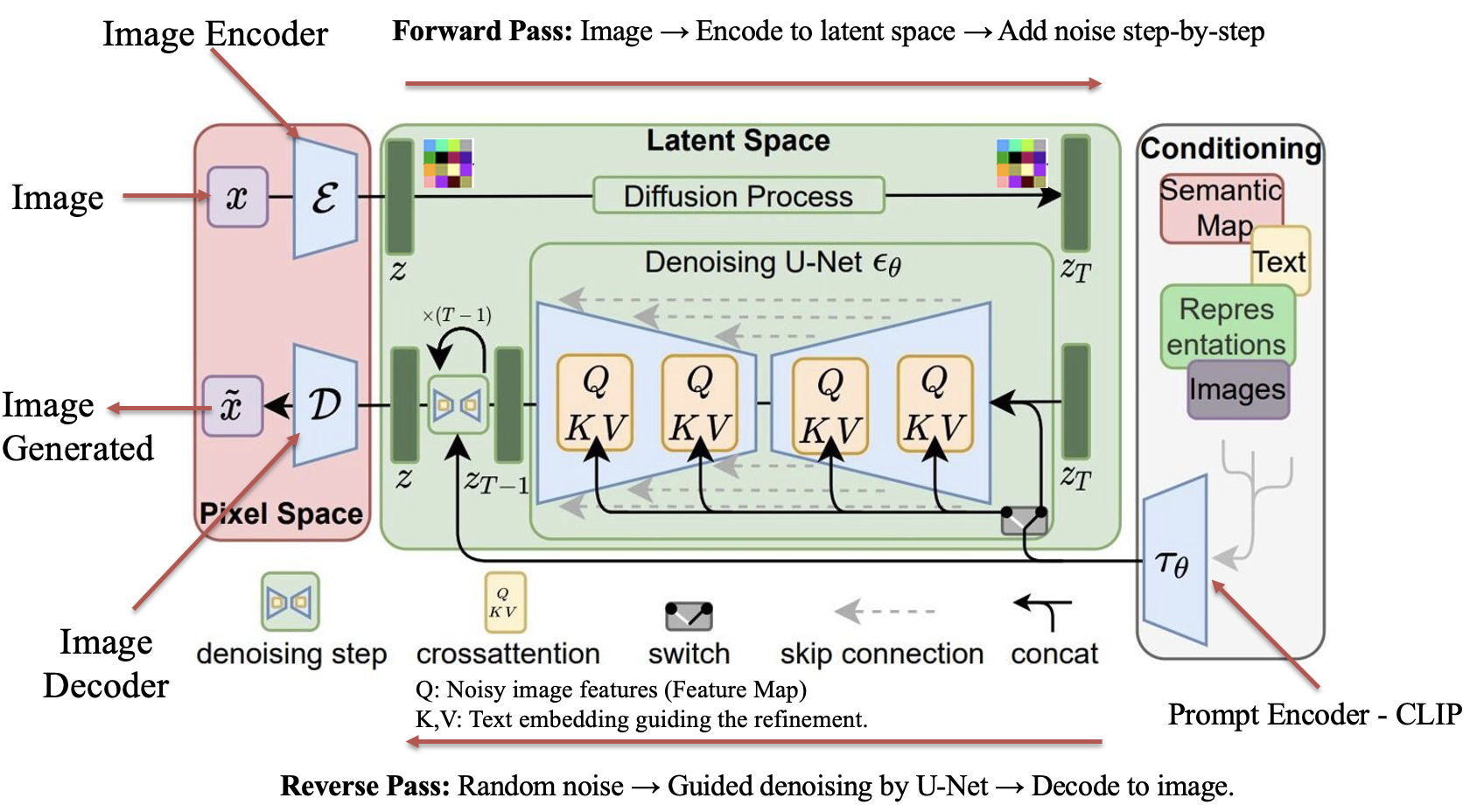}
    \caption{This figure illustrates the Stable Diffusion process for generating high-quality images from textual prompts. \cite{rombach2022high}.}
    \label{fig:stable_diffusion_pipeline}
\end{figure}

Stable Diffusion  operates in three main stages, as illustrated in Figure~\ref{fig:stable_diffusion_pipeline}:
\begin{enumerate}
    \item \textbf{Prompt Encoding:} The input text prompt is processed by a pre-trained text encoder, such as CLIP, to extract semantic embeddings that guide the image generation process. These embeddings are essential for conditioning the latent diffusion process.
    \item \textbf{Latent Space Diffusion:} The semantic embeddings are combined with noise in the latent space, where the diffusion process unfolds. A denoising U-Net iteratively refines the noisy latent representation into a structured latent image embedding. The U-Net uses cross-attention mechanisms to integrate text embeddings with image features, ensuring alignment between the prompt and the generated image.
    \item \textbf{Image Decoding:} The refined latent image representation is passed through a decoder (e.g., a pre-trained VAE decoder) to reconstruct the final high-resolution image in pixel space. The decoder ensures that the output adheres to the original semantics of the input text prompt.
\end{enumerate}

Figure~\ref{fig:stable_diffusion_pipeline} highlights the complete pipeline of Stable Diffusion, including the forward pass (adding noise to encode the image in latent space) and the reverse pass (iterative denoising conditioned on the text prompt). The integration of cross-attention mechanisms ensures that the output is not only semantically aligned but also visually coherent.

\begin{figure}[htbp]
    \centering
    \includegraphics[width=0.6\textwidth]{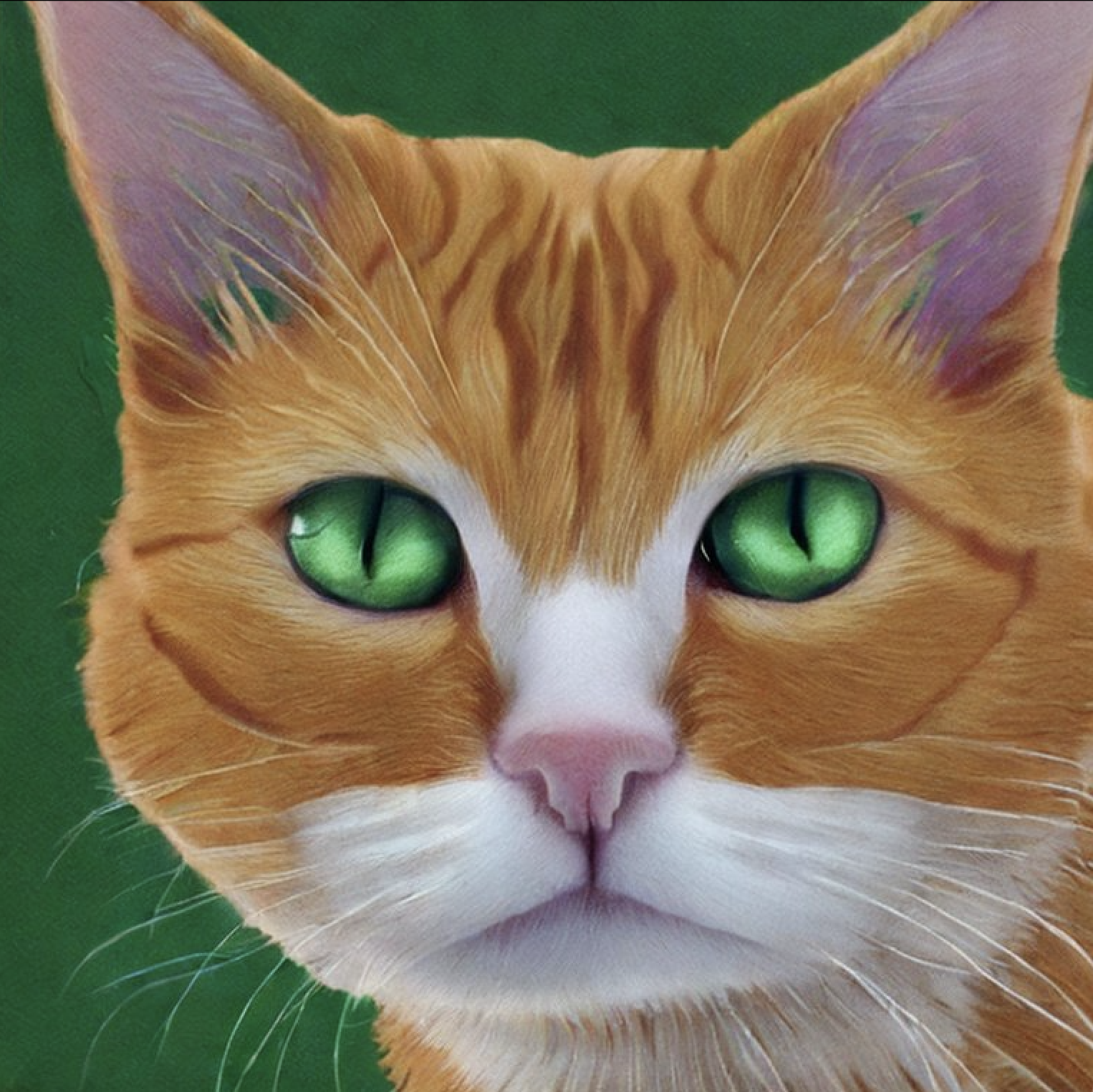}
    \caption{An image generated by Stable Diffusion for the prompt: "A photorealistic image of a fluffy orange tabby cat with green eyes." The model effectively captures realistic textures, lighting, and semantic details.}
    \label{fig:generated_cat}
\end{figure}

To illustrate the capabilities of Stable Diffusion, Figure~\ref{fig:generated_cat} showcases an image generated for the prompt: \textit{"A photorealistic image of a fluffy orange tabby cat with green eyes."}.

Stable Diffusion’s ability to efficiently generate high-resolution images, combined with its flexibility for creative and practical applications, makes it one of the most widely adopted diffusion-based frameworks for prompt-to-image generation.

\subsection{DALL·E: Autoregressive Frameworks}

DALL·E \cite{ramesh2022openai} is a pioneering framework for text-to-image generation that employs an autoregressive approach to synthesize images directly from textual descriptions. Unlike diffusion-based methods, DALL·E models the joint probability distribution of image and text tokens, enabling it to generate highly imaginative and diverse outputs in a sequential manner.

\begin{figure}[htbp]
    \centering
    \includegraphics[width=1.1\textwidth]{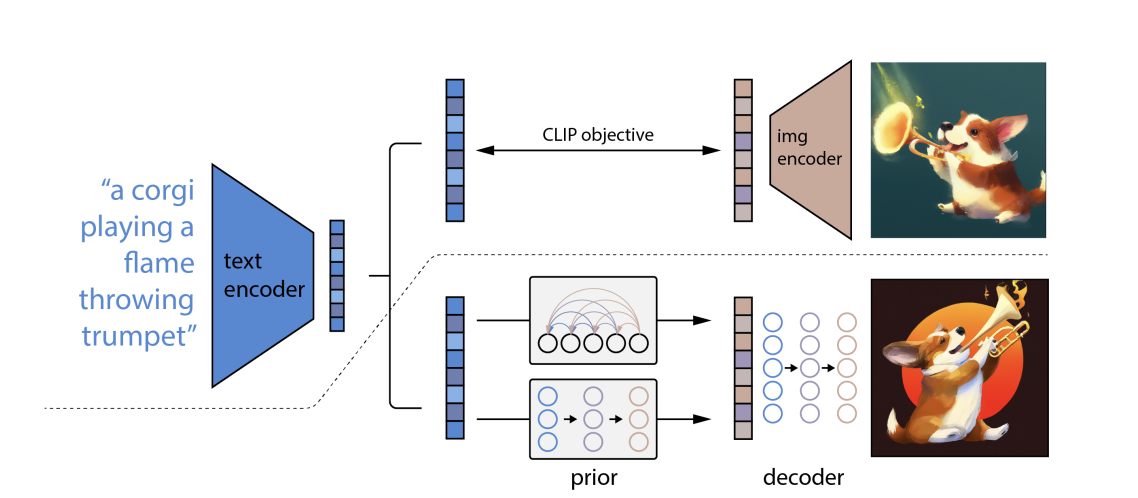}
    \caption{The DALL·E process for text-to-image generation: The text encoder processes the input prompt into embeddings, which are aligned with the image space using a CLIP-based objective. The prior predicts a latent image embedding, which is then decoded into a final image using the image decoder \cite{ramesh2022openai}.}
    \label{fig:dalle_process}
\end{figure}

\paragraph{How DALL·E Works:}
As illustrated in Figure~\ref{fig:dalle_process}, DALL·E operates through the following steps:

\begin{enumerate}
    \item \textbf{Text Encoding:} The input prompt (e.g., "a corgi playing a flame-throwing trumpet") is tokenized and processed by a pretrained text encoder, such as CLIP, to generate semantic embeddings. These embeddings serve as the initial context for the autoregressive generation process.
    \item \textbf{Prior Learning:} A prior model predicts a latent image embedding conditioned on the text embeddings. This step ensures that the generated image aligns semantically with the input prompt.
    \item \textbf{Autoregressive Generation:} Using a transformer-based architecture, DALL·E predicts the image tokens sequentially, conditioned on the latent embeddings and previously generated tokens.
    \item \textbf{Image Reconstruction:} The predicted image tokens are decoded into a complete image using the image decoder, ensuring the final output captures both the content and style specified in the prompt \cite{ramesh2022openai}.
\end{enumerate}

\begin{figure}[htbp]
    \centering
    \includegraphics[width=0.6\textwidth]{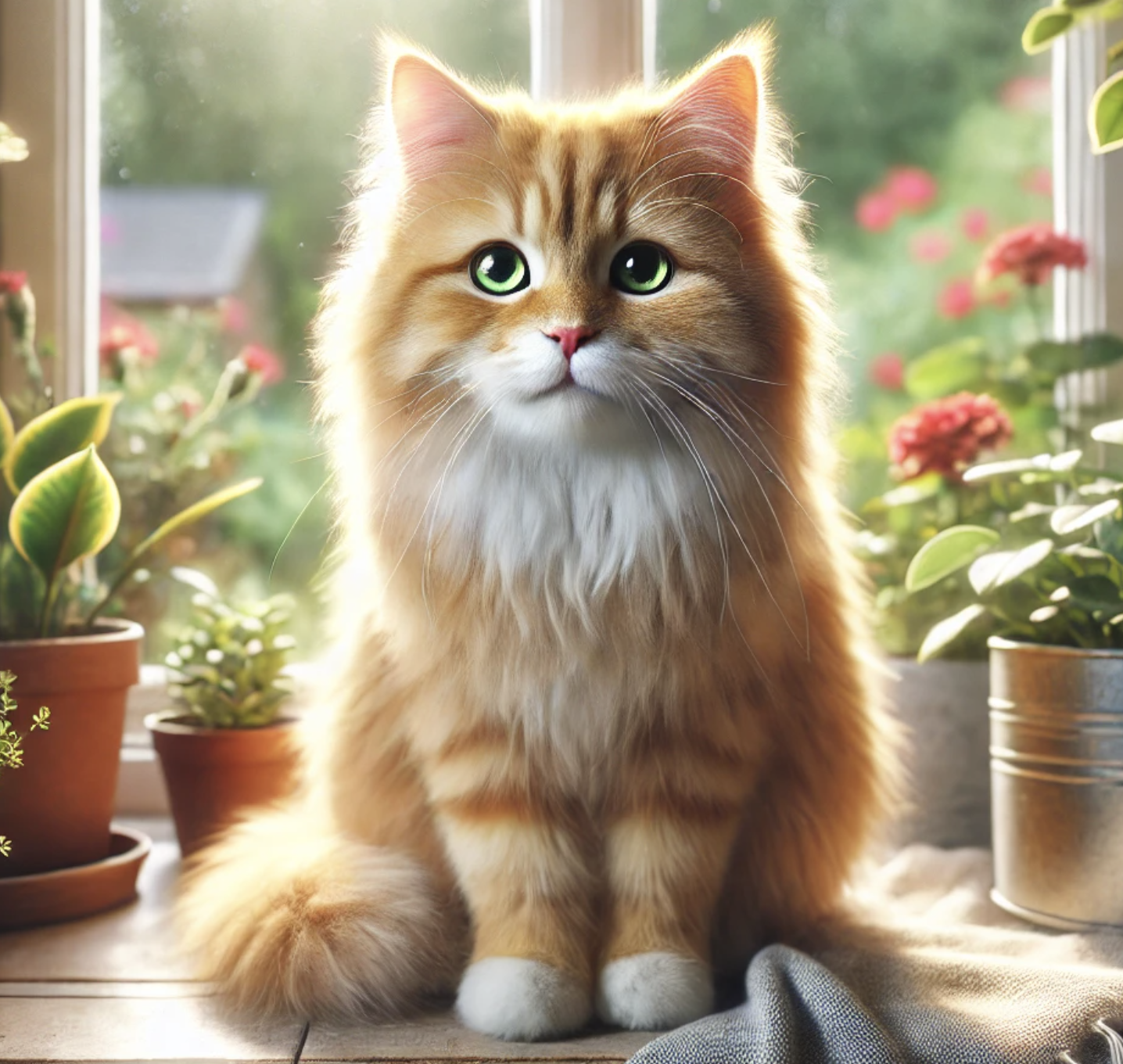}
    \caption{An image generated by DALL·E for the prompt: "A photorealistic image of a fluffy orange tabby cat with green eyes." The model effectively captures realistic details while retaining stylistic consistency. \cite{ramesh2022openai}}
    \label{fig:generated_cat_dalle}
\end{figure}

To illustrate DALL·E's capabilities, Figure~\ref{fig:generated_cat_dalle} shows an image generated for the prompt: \textit{"A photorealistic image of a fluffy orange tabby cat with green eyes."}.

DALL·E’s autoregressive approach enables it to generate highly creative and contextually rich outputs, making it particularly suitable for imaginative prompts. Its ability to handle abstract and complex instructions highlights its versatility in prompt-to-image generation.

\subsection{DeepSeek Janus-Pro}
DeepSeek Janus-Pro \cite{chen2025januspro} is a state-of-the-art multimodal framework designed to bridge the gap between visual understanding and visual generation. By leveraging a unified auto-regressive transformer, Janus-Pro facilitates seamless integration between image and text processing, enabling tasks such as image captioning, visual question answering, and text-to-image generation. The innovative decoupling of encoding processes for understanding and generation positions Janus-Pro as a flexible and robust solution for multimodal AI.

\begin{figure}[h]
    \centering
    \includegraphics[width=1\textwidth]{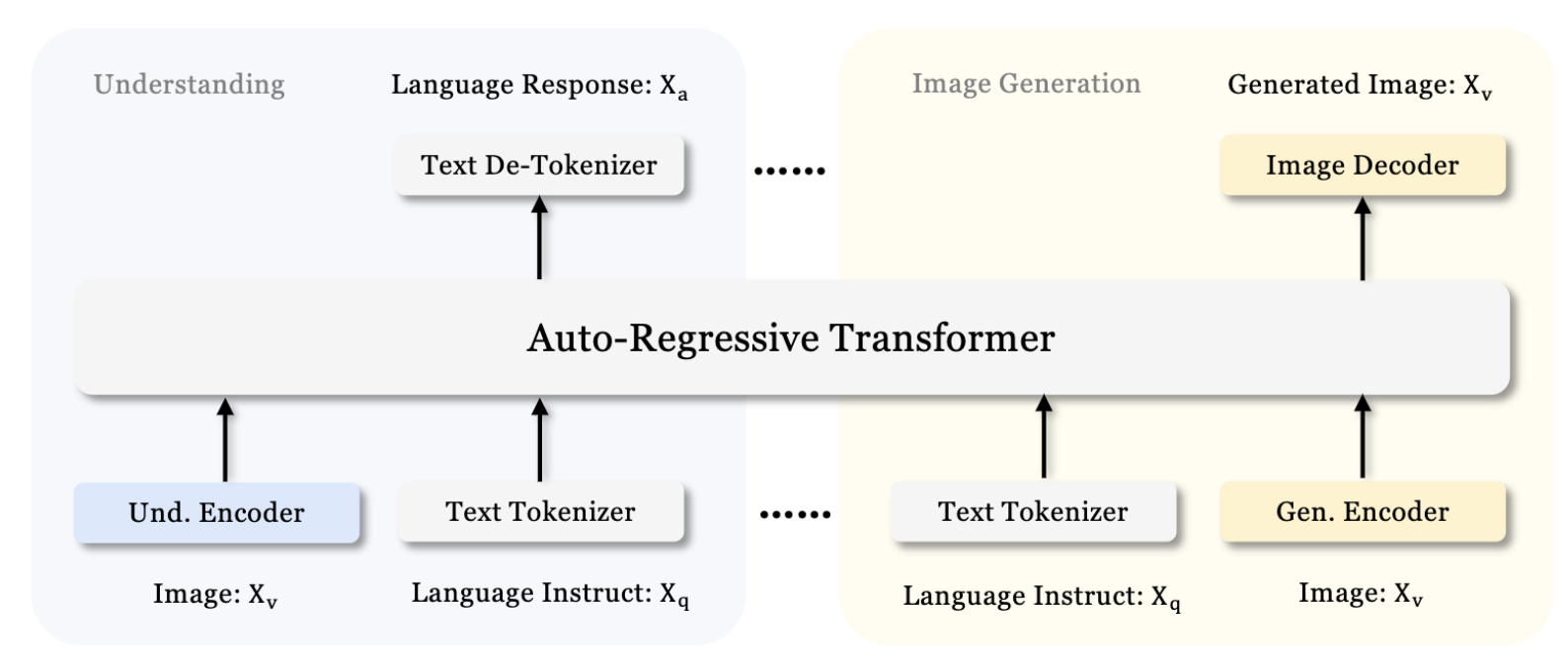} 
    \caption{Architecture of the Janus-Pro model. The framework decouples visual encoding into two modules: the Understanding Encoder for multimodal input comprehension and the Generation Encoder for output synthesis. Both are unified by an auto-regressive transformer, enabling seamless integration of visual understanding and generation \cite{chen2025januspro}.}
    \label{fig:janus_pro_architecture}
\end{figure}

\begin{figure}[H]
    \centering
    \includegraphics[width=0.8\textwidth]{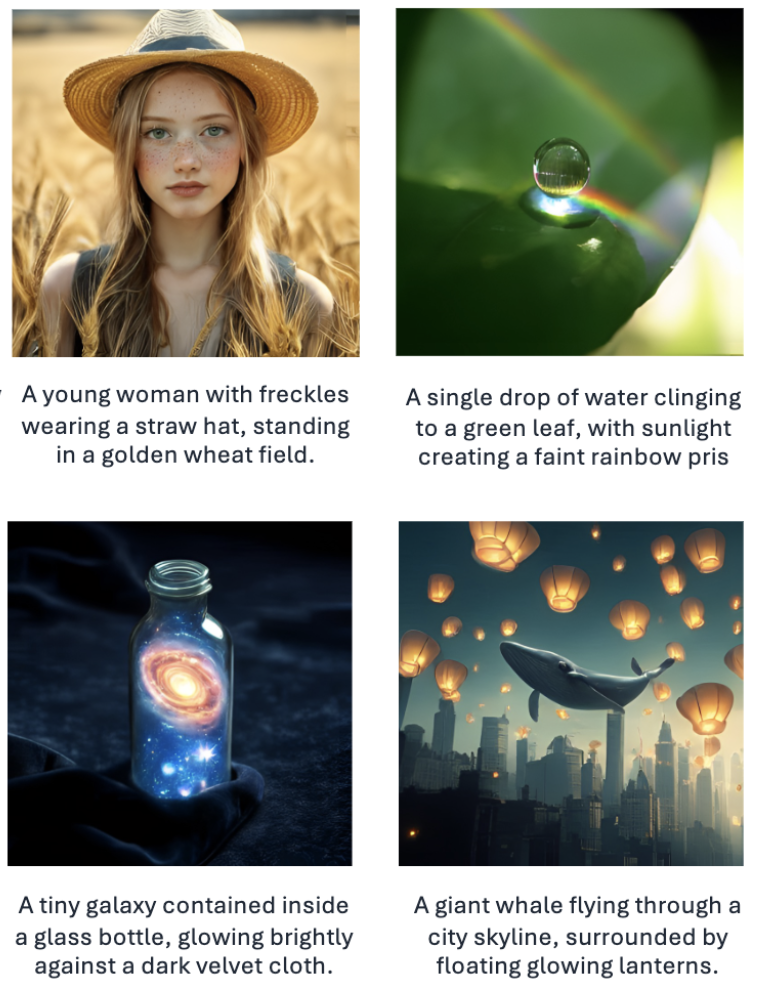} 
    \caption{Generated Images Using Prompt-to-Image Generation with DeepSeek Janus-Pro \cite{chen2025januspro}.}
    \label{image_janus_pro_architecture}
\end{figure}

\textbf{How DeepSeek Janus-Pro Works:}  
As illustrated in Figure~\ref{fig:janus_pro_architecture}, Janus-Pro operates through a dual-encoder framework. The model integrates an "Understanding Encoder" (Und. Encoder) for interpreting images and textual instructions, and a "Generation Encoder" (Gen. Encoder) for synthesizing visual outputs. Both components are mediated by a shared auto-regressive transformer, which unifies the understanding and generation processes in a scalable and cohesive manner. The workflow is as follows:
\begin{itemize}
    \item \textbf{Step 1: Visual Understanding}  
    The input image ($X_v$) is processed by the Understanding Encoder, extracting meaningful representations. Simultaneously, textual instructions ($X_q$) are tokenized and encoded for cross-modal alignment.
    \item \textbf{Step 2: Unified Processing}  
    Encoded representations from both modalities are fed into the auto-regressive transformer, which generates a unified latent representation for downstream tasks.
    \item \textbf{Step 3: Visual Generation}  
    For text-to-image tasks, the Generation Encoder combines the unified latent representation with tokenized textual instructions to produce a structured latent visual output. This is then passed through the Image Decoder to synthesize the final image.
    \item \textbf{Step 4: Language Response (if applicable)}  
    For tasks requiring textual output, such as visual question answering, the latent representation is decoded into text tokens, de-tokenized to produce the language response.
\end{itemize}

Figure~\ref{image_janus_pro_architecture} illustrates the generated images using prompt-to-image generation with DeepSeek Janus-Pro. The results showcase the model's ability to produce high-quality, detailed, and realistic visuals, effectively aligning the generated outputs with the given textual prompts. \\

\textbf{Key Features of DeepSeek Janus-Pro}  
\begin{enumerate}
    \item \textbf{Decoupled Encoding Architecture:}  
    The model separates encoding for understanding and generation, allowing optimized processing paths for each modality while maintaining a shared transformer backbone.
    \item \textbf{Unified Auto-Regressive Transformer:}  
    A central transformer processes multimodal inputs cohesively, ensuring contextual alignment between text and image representations.
    \item \textbf{Flexibility Across Modalities:}  
    Janus-Pro supports diverse tasks, including image captioning, visual question answering, and text-to-image synthesis, making it a versatile framework for multimodal applications.
    \item \textbf{Scalable Design:}  
    The modular architecture allows Janus-Pro to scale effectively for tasks of varying complexity, from simple captioning to high-resolution image generation.
\end{enumerate}

\subsection{Applications of Prompt-to-Image Generation}
\begin{enumerate}

\item \textbf{Marketing and Advertising:} \begin{itemize} \item \textbf{Use Case:} Generating tailored visual content for brand campaigns and product promotions. \item \textbf{Process:} Fine-tune a multimodal model like CLIP using image data paired with text descriptions. Use prompts such as “a minimalist coffee mug in a modern kitchen” to generate visuals aligned with campaign goals. \item \textbf{Impact:} Reduces creative production time and costs, enabling personalized and dynamic campaigns that enhance customer engagement. \end{itemize}

\item \textbf{E-Commerce:} \begin{itemize} \item \textbf{Use Case:} Creating dynamic product visuals to help customers visualize items in real-world contexts. \item \textbf{Process:} Train a text-to-image generation model like Stable Diffusion using product images and annotated text. Generate visuals such as “a red sofa in a cozy living room” to present products more effectively. \item \textbf{Impact:} Improves customer confidence by offering personalized previews, reducing return rates, and increasing sales. \end{itemize}

\item \textbf{Architecture and Real Estate:} \begin{itemize} \item \textbf{Use Case:} Visualizing designs, layouts, and property renovations for clients and stakeholders. \item \textbf{Process:} Train text-to-image frameworks using architectural blueprints and design annotations. Generate realistic layouts, such as “a modern living room with natural light and wooden flooring,” to illustrate concepts. \item \textbf{Impact:} Speeds up the design process, reduces costs for renderings, and enhances client engagement by providing realistic previews. \end{itemize}

\end{enumerate}

These applications demonstrate the versatility and impact of prompt-to-image generation across a wide range of industries, showcasing its potential to enhance productivity, creativity, and user engagement.

\section{Conditional Inputs for Image Generation}
\label{Conditional}
Conditional inputs provide a powerful mechanism for guiding image generation processes by incorporating user-defined constraints, such as textual prompts, labels, sketches, depth maps, or other structural data. Unlike traditional generation methods relying solely on noisy vectors or latent embeddings, conditional models align generated outputs with specific semantic, spatial, and stylistic requirements. By leveraging multimodal inputs, these models enable fine-grained control and greater customization of generated visuals.

\subsection{Mixed Inputs: Combining Text and Visual Guidance}

The integration of textual and visual inputs provides a versatile approach to generating images that adhere to both semantic intent and structural constraints. Models in this category combine textual prompts with structural guidance, such as edge maps, segmentation maps, or depth cues, ensuring coherence across modalities.

\paragraph{Prominent Frameworks:}
\begin{itemize}
    \item \textbf{ControlNet:} A diffusion-based framework that integrates visual cues, such as edge detection, depth maps, and pose keypoints, with text prompts to achieve precise alignment between structure and semantics \cite{zhang2023controlnet}.
    \item \textbf{T2I-Adapter:} Builds on text-to-image diffusion models by incorporating adapters for structural input signals, such as segmentation maps or layouts, enabling fine control over output characteristics \cite{mou2023t2iadapter}.
    \item \textbf{InstructPix2Pix:} Specializes in editing existing images using textual instructions combined with the original image. For example, modifying a living room photo with the instruction "make the walls blue" produces a realistic, updated image \cite{brooks2022instructpix2pix}.
    \item \textbf{DreamBooth:} Focuses on personalization by combining reference images with text prompts to fine-tune models for generating subject-specific outputs \cite{ruiz2022dreambooth}.
    \item \textbf{Textual Inversion:} Embeds reference image features into the latent space of diffusion models, allowing users to generate stylistically consistent visuals aligned with text descriptions \cite{gal2022textualinversion}.
\end{itemize}

\subsection{ControlNet and Advanced Derivatives}

ControlNet \cite{zhang2023controlnet} extends text-to-image diffusion models, such as Stable Diffusion, by integrating additional control signals like edge maps, depth maps, or pose keypoints. These inputs enable precise structural guidance during the image generation process, complementing the semantic cues provided by text prompts.

ControlNet works by attaching control modules to a pre-trained diffusion model. These modules process structural inputs in parallel with text embeddings, ensuring alignment between visual constraints and semantic intent. The key steps are as follows:
\begin{enumerate}
    \item \textbf{Input Encoding:} The text prompt is encoded using a pretrained text encoder (e.g., CLIP \ref{Multimodal_encoder}), while structural data is processed through ControlNet’s modules.
    \item \textbf{Latent Alignment:} The text and structural embeddings are aligned in a shared latent space.
    \item \textbf{Guided Diffusion:} The diffusion model uses both embeddings to refine the output, adhering to both semantic and structural inputs.
\end{enumerate}

\begin{figure}[htbp]
    \centering
    \includegraphics[width=\textwidth]{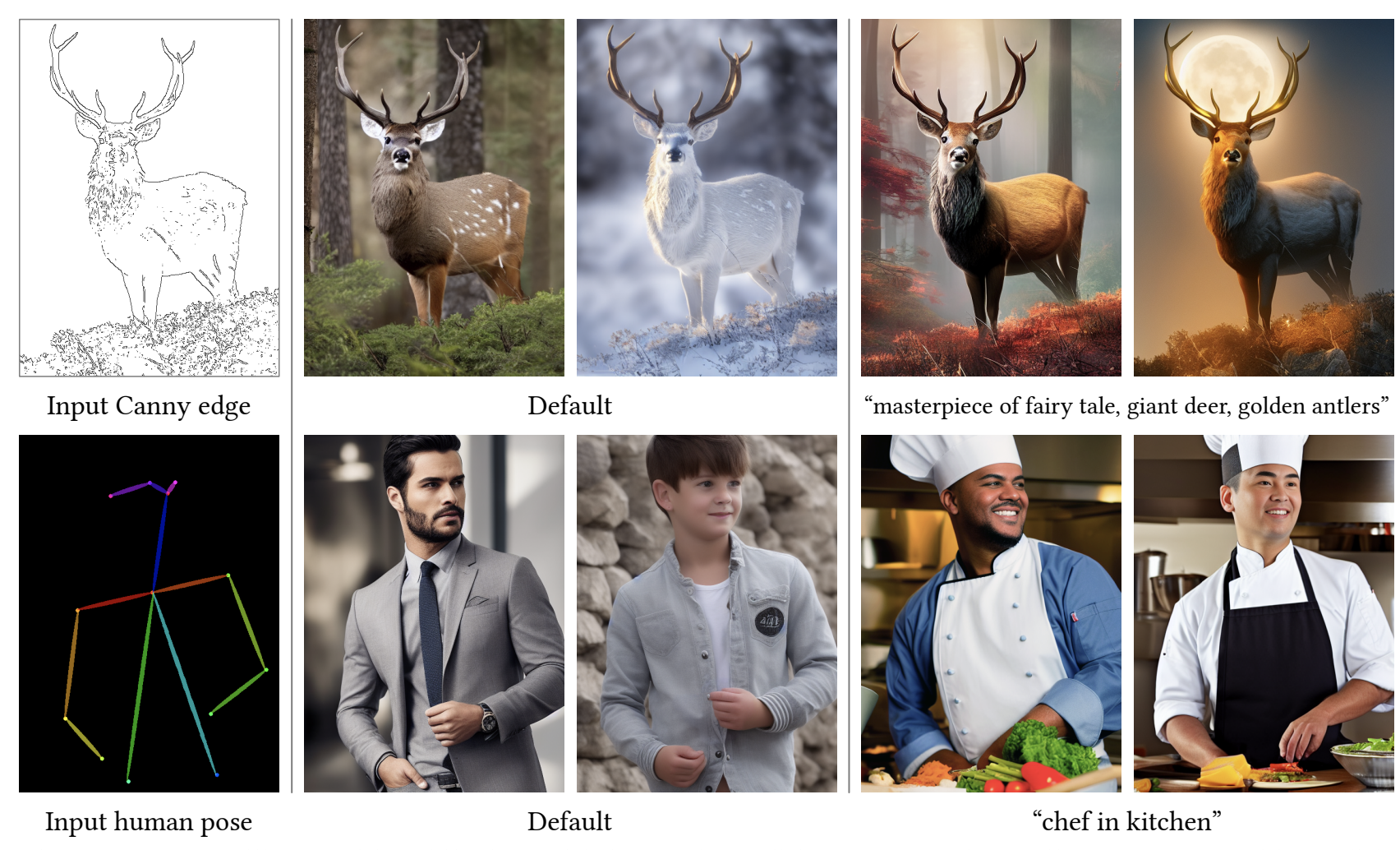}
    \caption{Examples of image generation using ControlNet. Top row: Input Canny edge maps guide the structural layout of deer images, with default results and further customization based on text prompts such as "masterpiece of fairy tale, giant deer, golden antlers" and "quaint city Galic." Bottom row: Input human poses are used to control the generation of realistic human figures, including customized outputs based on prompts like "chef in kitchen." ControlNet enables precise control over generated visuals by combining structural inputs with text prompts \cite{zhang2023controlnet}.}
    \label{fig:controlnet_examples_updated}
\end{figure}

ControlNet integrates seamlessly with Stable Diffusion by leveraging its robust latent space and denoising capabilities. For instance, combining a sketch with the prompt "a majestic deer in the forest" ensures that the output is both structurally accurate and semantically aligned. Figure~\ref{fig:controlnet_examples_updated} demonstrates examples of images generated using ControlNet with Canny edge maps and human pose inputs.

\paragraph{Key Features of ControlNet}
\begin{itemize}
    \item \textbf{Edge Detection:} Accepts edge maps to define contours and structural layouts, ensuring adherence to specific shapes or boundaries in the output.
    \item \textbf{Depth Maps:} Utilizes depth information to guide the 3D structure of generated visuals, ideal for AR/VR and 3D rendering applications.
    \item \textbf{Pose Keypoints:} Allows precise control over character positioning by conditioning outputs on human or object poses.
    \item \textbf{Semantic Segmentation:} Uses segmentation maps to assign specific categories (e.g., sky, buildings, vegetation) to different regions of the generated image.
    \item \textbf{Style and Texture Control:} Enables users to define specific textures or styles, producing outputs that are both structurally and stylistically coherent.
\end{itemize}

\paragraph{Derivatives of ControlNet:}
\begin{itemize}
    \item \textbf{T2I-Adapter:} Expands upon ControlNet by integrating specialized adapters that provide additional control over layout, depth, and style \cite{mou2023t2iadapter}.
    \item \textbf{PoseGuidedGAN:} Combines pose estimation with GAN-based architectures for pose-controlled image generation, particularly in character design and animation \cite{ma2017poseguidedgan}.
    \item \textbf{Compositional Diffusion:} Integrates multiple conditional inputs (e.g., text, edge maps, and reference images) to generate visually and semantically aligned outputs for complex scenes.
\end{itemize}

\newpage

\subsection{Applications of Conditional Inputs}

The integration of structural, semantic, and stylistic guidance in frameworks like ControlNet enables impactful use cases in various industries. Below are the top three industries with the highest impact:

\paragraph{Creative Design and Art}

\begin{itemize}
    \item \textbf{Use Case:} Tailored artwork creation for advertisements, media, or entertainment.
    \item \textbf{Process:} Combine sketches or segmentation maps with text prompts like “a futuristic city skyline at sunset” to guide image generation.
    \item \textbf{Impact:} Streamlines the creative workflow, reduces production time, and enables designers to visualize complex ideas with precision.
\end{itemize}

\paragraph{Healthcare}
\begin{itemize}
    \item \textbf{Use Case:} Enhanced medical imaging for diagnostics and training.
    \item \textbf{Process:} Input segmentation maps or CT/MRI scans into conditional frameworks to generate detailed visuals that highlight specific anomalies or structures.
    \item \textbf{Impact:} Improves diagnostic accuracy, supports medical training with high-fidelity visuals, and reduces reliance on real patient data for research.
\end{itemize}

\paragraph{E-Commerce and Retail}
\begin{itemize}
    \item \textbf{Use Case:} Realistic product visualizations for online platforms.
    \item \textbf{Process:} Use a depth map or sketch paired with a prompt like “a modern armchair in a minimalist room” to produce visually compelling product representations.
    \item \textbf{Impact:} Enhances customer engagement, reduces costs associated with photoshoots, and accelerates product listings with dynamic visual options.
\end{itemize}
The integration of structural, semantic, and stylistic guidance in conditional input frameworks like ControlNet and its derivatives promises to redefine creative and technical workflows, enabling unprecedented levels of precision and customization.

\section{Challenges in Image Generation}

\label{ethical}
Despite its remarkable advancements, image generation through generative AI faces several critical challenges that hinder its widespread adoption and reliability. These challenges span technical, ethical, and operational domains, emphasizing the need for continued research and innovation to ensure equitable and efficient deployment.

\paragraph{Bias and Fairness:} Generative models often inherit biases from their training data, reflecting societal, cultural, or demographic disparities. For instance, a text-to-image model may underrepresent certain ethnicities, genders, or cultural symbols in its outputs, leading to skewed or stereotypical results. Addressing bias requires careful dataset curation, enhanced training techniques, and post-generation fairness auditing to mitigate these issues while promoting inclusivity and diversity.

\paragraph{Computational Cost and Efficiency:} Training and deploying state-of-the-art generative models, particularly diffusion-based frameworks or GANs, demand significant computational resources. The high cost of GPUs, energy consumption, and extended training times pose barriers for small-scale researchers and businesses. Strategies like model optimization, transfer learning, and efficient architectures are essential to reduce these resource demands while maintaining output quality.

\paragraph{Unfair Representation and Content Misuse:} Generative models have raised concerns about the potential misuse of generated content, including deepfakes, misinformation, and unauthorized reproductions of copyrighted materials. These challenges necessitate robust policies, watermarking techniques, and model safeguards to prevent malicious applications while ensuring ethical use.

\paragraph{Alignment with User Intent:} Aligning generated outputs with user-defined intents remains a complex challenge. Ambiguous prompts, conflicting constraints, or limitations in multimodal alignment can lead to outputs that deviate from user expectations. Enhanced training techniques, better prompt engineering, and reinforcement learning from human feedback (RLHF) are promising solutions to improve alignment accuracy and consistency.

\paragraph{Ethical and Societal Implications:} The deployment of generative AI raises broader ethical and societal concerns, such as the displacement of creative professionals, the perpetuation of harmful stereotypes, and the potential erosion of trust in visual content. Addressing these issues requires a multidisciplinary approach, incorporating perspectives from policy, ethics, and technical development to ensure responsible innovation.

\section{Conclusion and Future Directions}
\label{conclusion}

This work has presented a comprehensive evaluation of generative AI frameworks for vision, proposing a structured categorization of image generation techniques based on input modalities. By analyzing state-of-the-art methodologies, including noisy vector-based models, latent space representations, and conditional frameworks, we have highlighted their foundational principles, practical applications, and underlying challenges. The integration of text-to-image models like DALL·E and Stable Diffusion, as well as advanced frameworks such as ControlNet, demonstrates the transformative potential of generative AI in creating contextually rich and visually coherent content. This categorization aims to provide clarity for researchers, application developers, and industry practitioners seeking to navigate the evolving landscape of vision-focused generative AI.

Despite these advancements, the field faces persistent challenges, including computational inefficiencies, bias mitigation, and alignment of generated outputs with user intent. Addressing these issues will require innovation in algorithmic design, ethical governance, and operational scalability.

To ensure the continued impact of generative AI in vision, future research and development should prioritize the following areas:

\paragraph{Advancements in Multimodal Alignment:} Enhanced multimodal models that unify vision, language, and spatial data will pave the way for improved cross-domain understanding. Frameworks capable of seamless integration of diverse inputs, such as text and structural guidance, will unlock new levels of precision and flexibility in content generation.

\paragraph{Operational Scalability:} Reducing the computational cost and energy consumption of training and deploying large-scale generative models is crucial. Techniques like model compression, efficient latent space representations, and hardware optimization can make generative AI more accessible to smaller organizations and developers.

\paragraph{Bias and Ethical Considerations:} Developing frameworks that identify and mitigate biases in training data will ensure fair representation across demographics and cultural contexts. Future models must incorporate fairness evaluation metrics and ethical guidelines to promote inclusivity and prevent harmful outputs.

\paragraph{Interactive and Real-Time Generation:} Real-time generative AI systems, particularly in AR/VR applications and gaming, represent a significant opportunity. Advances in temporal consistency for video generation and real-time adaptation of outputs will enhance user engagement and operational utility.

\paragraph{Industry-Specific Applications:} Tailored solutions for industries such as healthcare, manufacturing, and retail will drive deeper integration of generative AI into operational workflows. For instance, models optimized for medical imaging or product customization could offer domain-specific advantages.

\paragraph{Transparency and Interpretability:} Ensuring that generated content aligns with user expectations requires models that provide clear explanations of their decisions and outputs. Enhancing interpretability will build trust and facilitate adoption in high-stakes applications.

\paragraph{Agentic Systems for Vision:} Vertical AI agents \cite{bousetouane2025agenticsystemsguidetransforming}\cite{bousetouane2025physicalaiagentsintegrating}, leveraging multimodal understanding, will play a pivotal role in vision-driven applications. By integrating vision, language, and spatial awareness, these agents can autonomously perceive, reason, and act within dynamic environments. Their ability to adapt and interact with surroundings will enhance applications in robotics, autonomous systems, manufacturing, and healthcare. As they evolve, their capacity for contextual reasoning and real-time decision-making will drive transformative advancements across industries.

By tackling existing challenges and driving innovation, generative AI for vision can continue to advance, shaping robust, ethical, and impactful solutions that address real-world needs and unlock new possibilities across industries.

\bibliographystyle{plain} 
\bibliography{science_template} 

\end{document}